\documentclass[conference]{IEEEtran}
\IEEEoverridecommandlockouts
\usepackage{cite}
\usepackage{amsmath,amssymb,amsfonts}
\usepackage{algorithmic}
\usepackage{graphicx}
\usepackage{textcomp}
\usepackage{xcolor}
\usepackage{hyperref}
\def\BibTeX{{\rm B\kern-.05em{\sc i\kern-.025em b}\kern-.08em
    T\kern-.1667em\lower.7ex\hbox{E}\kern-.125emX}}
\begin{document}

\title{An Effective Image Copy-Move Forgery Detection Using Entropy Information}

\author{\IEEEauthorblockN{ Li Jiang}
\IEEEauthorblockA{\textit{School of Electrical and Information Engineering} \\
\textit{Zhengzhou University}\\
Zhengzhou, China \\
ieljiang@zzu.edu.cn}
~\\
\and
\IEEEauthorblockN{ Zhaowei Lu*}
\IEEEauthorblockA{\textit{School of Electrical and Information Engineering} \\
\textit{Zhengzhou University}\\
Zhengzhou, China \\
luzhaoweizzu@163.com}
*Corresponding author}

\maketitle

\begin{abstract}
Image forensics has become increasingly crucial in our daily lives. Among various types of forgeries, copy-move forgery detection has received considerable attention within the academic community. Keypoint-based algorithms, particularly those based on Scale Invariant Feature Transform, have achieved promising outcomes. However, most of keypoint detection algorithms failed to generate sufficient matches when tampered patches were occurred in smooth areas, leading to insufficient matches. Therefore, this paper introduces entropy images to determine the coordinates and scales of keypoints based on Scale Invariant Feature Transform detector, which make the pre-processing more suitable for solving the above problems. Furthermore, an overlapped entropy level clustering algorithm is developed to mitigate the increased matching complexity caused by the non-ideal distribution of gray values in keypoints. Experimental results demonstrate that our algorithm achieves a good balance between performance and time efficiency.
\end{abstract}

\begin{IEEEkeywords}
Image forensics, copy-move forgery detection, Scale Invariant Feature Transform, entropy level clustering
\end{IEEEkeywords}

\section{Introduction}\label{Sec1}
%However, not all types of manipulations are concerned. In practical applications, people often focus on forgeries that cause semantic changes. In this scenario, copy-move and splicing have gained significant attention in the academic community. Currently, digital image forgery detection techniques can be classified into active and passive methods. Active techniques, such as digital watermarking and signatures, aim to detect tampering by verifying the integrity of pre-embedded prior information in the image. However, these techniques may affect the overall quality of the image. In contrast, passive techniques solely rely on the content information within the image itself and do not alter the original data, making them more widely applicable.
With the advancement of multimedia technology, the quality of digital image forgeries has improved significantly. Simultaneously, the cost associated with such forgeries has decreased. Consequently, it has become increasingly challenging for people to trust the authenticity of images, unlike several decades ago. Among the various manipulations, copy-move operation is particularly challenging due to its inherent similarity. Conventionally, Copy-Move Forgery Detection (CMFD) can be mainly divided into keypoint-based algorithms \cite{1,2,3,4,5,14,15,17} and block-based algorithms \cite{6,7}. Currently, with the rise of deep learning developed, deep-based algorithms \cite{8,9,10,11,12} have gradually been applied in this field. The main differences between conventional algorithms and deep-based algorithms are as follows:
\begin{itemize}
	\item Deep-based algorithms outperform conventional on low-resolution images, but they usually rely on downsampling to detect high-resolution images; Conventional algorithms can detect all image sizes, but take longer with high-resolution images and rely on hand-crafted features for computer vision tasks.
	\item Deep-based algorithms still lack interpretability, while conventional algorithms are known for their good interpretability.
	\item Deep-based algorithms can be used to distinguish between source and target regions, while conventional algorithms struggle with this task.
\end{itemize}

Although deep-based algorithms have achieved great results, downsampling strategies for processing high-resolution images may lose important information, resulting in detection failure. Therefore, it is still necessary to continue to develop the conventional CMFD algorithm. This paper conducted research on the popular keypoint-based algorithms in CMFD. In existing works, detecting keypoints in gray images are the most commonly applied methods \cite{1,3,4,5,14}. However, gray images primarily represent brightness, making conventional keypoint detection algorithms less effective in regions with low texture. To address these issues, this paper proposed an effective CMFD algorithm, which mainly includes:
\begin{itemize}
	\item This paper introduces entropy images to determine the coordinates and scales of keypoints based on Scale Invariant Feature Transform (SIFT) detector \cite{13}. Since SIFT features represent the gradient information of the grayscale, the keypoints redefine orientation and extraction feature in gray image, which make matching stage more accuracy.
	\item  This paper develops an overlapped entropy level clustering algorithm, which greatly reduces computation complexity caused by non-ideal grayscale distribution of keypoints.
\end{itemize}

The remainder of this paper is organized as follows. Related work is reported in section \ref{Sec2}. Our proposed method is introduced in section \ref{Sec3}. Experimental results are given in section \ref{Sec4}, and section \ref{Sec5} draws the conclusion.

\section{Related technique}\label{Sec2}
\subsection{Review Of The SIFT}\label{Sec2_1}
Classical SIFT consists of the following four steps \cite{13}:
\begin{itemize}
	\item determining candidate points in Difference of Gaussian (DoG) space.
	\item selecting keypoints using contrast threshold $ T_{con} $.
	\item calculating the dominant orientation.
	\item generating the feature descriptor.	
\end{itemize}

Suppose there is a gray image $ I_{gray} $, the candidate points are extracted by searching local extrema within $ 3 \times 3 \times 3 $ DoG space regions. The DoG image $ D $ at scale $ \sigma $ is given by:
\begin{equation}
	\label{Eq1}
	D(x,y,\sigma ) = L(x,y,k\sigma ) - L(x,y,\sigma )
\end{equation} 
Here, $ k $ is a predefined constant,  $ (x,y) $ represents coordinates, and $ L $ means the gaussian image of $ I_{gray} $. Mathematically, $ L $ can be defined as:
\begin{equation}
	\label{Eq2}
	L(x,y,\sigma ) = {I_{gray}}(x,y) \otimes {\mathop{\rm G}\nolimits} (x,y,\sigma )
\end{equation} 
Here, $ \rm G $ and $ \otimes $ represent the Gaussian kernel and  convolution operation, respectively. By applying Equation (\ref{Eq1}), multiple DOG spaces are generated at various scales.

Subsequently, the contrast values $ |D(\textbf{x})| $ in the candidate points are calculated:
\begin{equation}
	\label{Eq3}
	|D(\textbf{x})| = |D + \frac{1}{2}{(\frac{{\partial D}}{{\partial {\textbf{x}}}})^T}{\textbf{x}}|
\end{equation} 
Here, $ {\textbf{x}} $ represents the coordinates of the candidate points. If the contrast value of candidate point is greater than the contrast threshold $ T_{con} $, the candidate point will be selected as a keypoint.

Finally, the keypoints $ \bf{KP} $ and features $ \bf{F} $ will be represented as:
\begin{equation}
	\label{Eq4}
	{\bf{KP}} = \{ \bf{x},\bf{y},\bf{\sigma},\bf{\theta}\}
\end{equation} 
\begin{equation}
	\label{Eq5}
	{\bf{F}} = \left\{ {{{\bf{f_1}}},{{\bf{f_2}}}, \cdots ,{{\bf{f_{128}}}}} \right\}
\end{equation} 
Here, $ \bf{F} $ represents 128-dimensional features, and $ \bf{\theta} $ represents dominant orientation:
\begin{equation}
	\label{Eq6}
	\theta  = \arctan (\frac{{L(x,y + 1) - L(x,y - 1)}}{{L(x + 1,y) - L(x - 1,y)}})
\end{equation}

\subsection{Detect Smooth Or Small Tampering}\label{Sec2_2}
\begin{figure}[ht]
	\centering
	\begin{tabular}{cc}
		\includegraphics[width=0.45\linewidth]{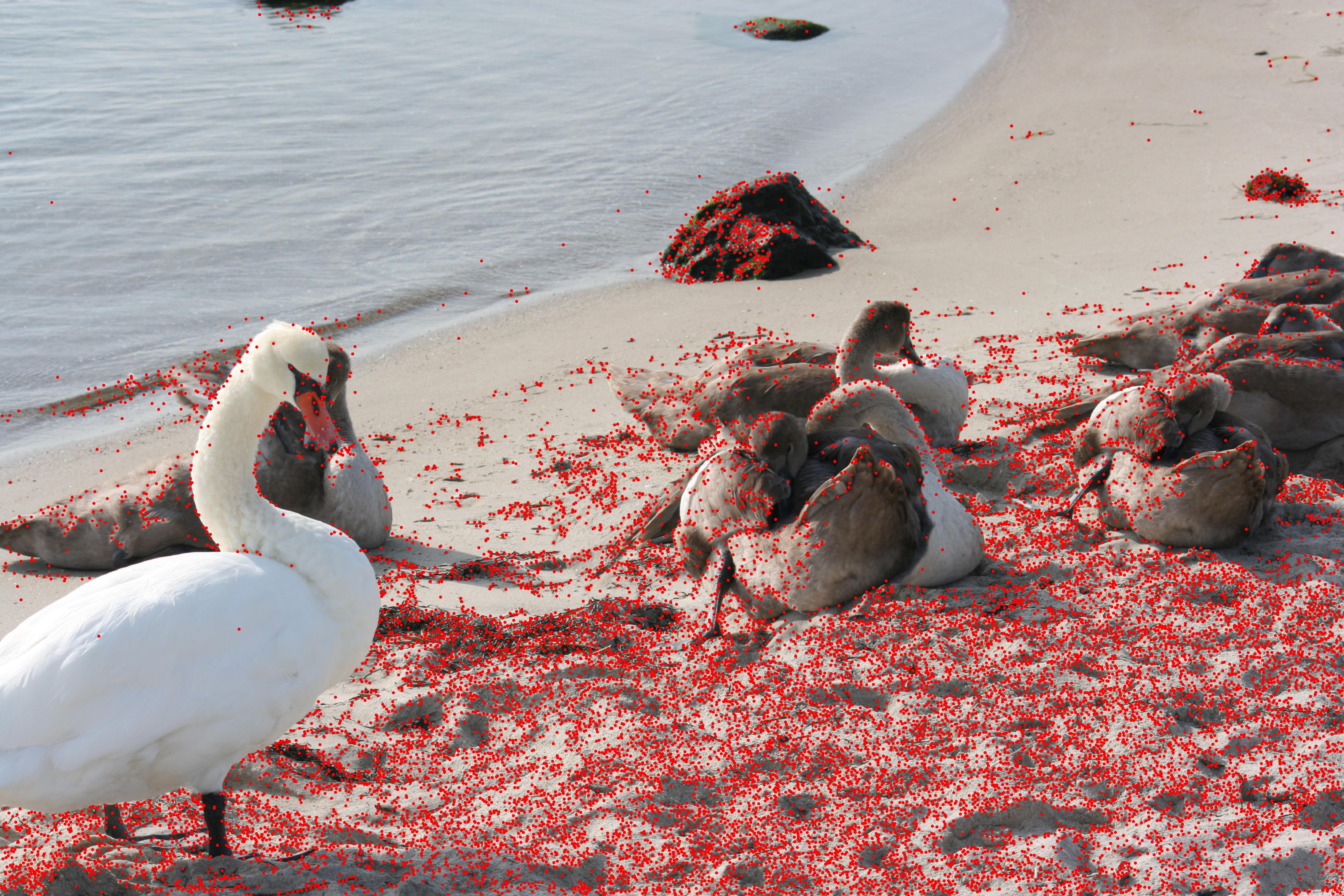} &
		\includegraphics[width=0.45\linewidth]{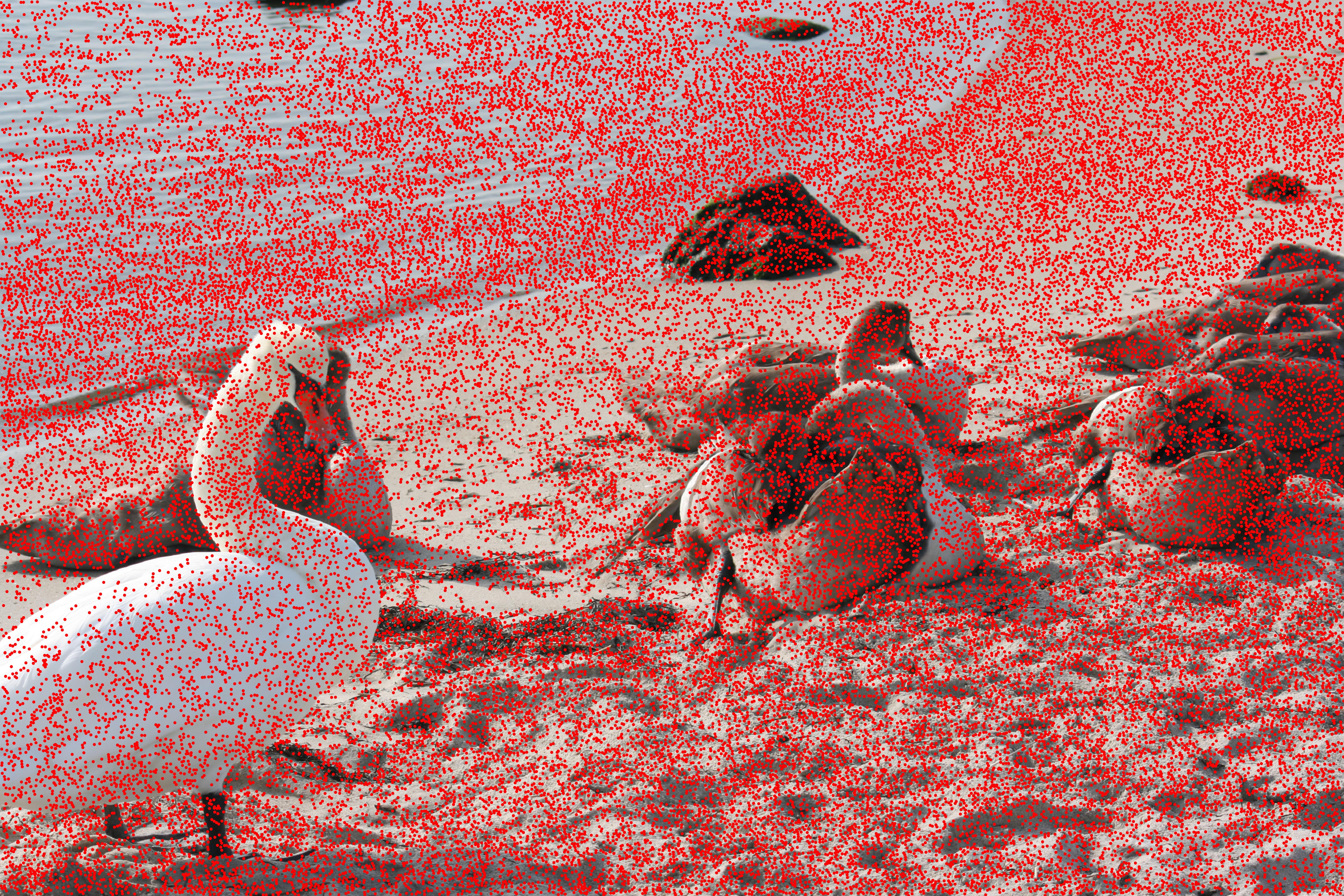} \\
		(a) & (b)
	\end{tabular}
	\caption{Keypoint detection by (a) classical contrast threshold; (b) reducing contrast threshold.}
	\label{Fig1}
\end{figure}

In recent years, the most important challenge facing CMFD is how to accurately detect tampering in small or smooth areas. In order to solve the above two problems, the SIFT algorithm has two strategies. The first strategy is reducing the contrast threshold. Compared with the results of the classical SIFT algorithm shown in Fig. \ref{Fig1} (a), the strategy of reducing the contrast threshold shown in Fig. \ref{Fig1} (b) can significantly increase the number of keypoints in the smooth region. The second strategy is image upsampling. This strategy aims to increase the number of keypoints per unit region of the original image. Through a reasonable number of upsampling, we can obtain sufficient keypoints for tampering localization. However, these two strategies are often only used for gray images.

\begin{figure*}[ht]
	\centering
	\begin{tabular}{ccccc}
		\includegraphics[width=0.18\linewidth]{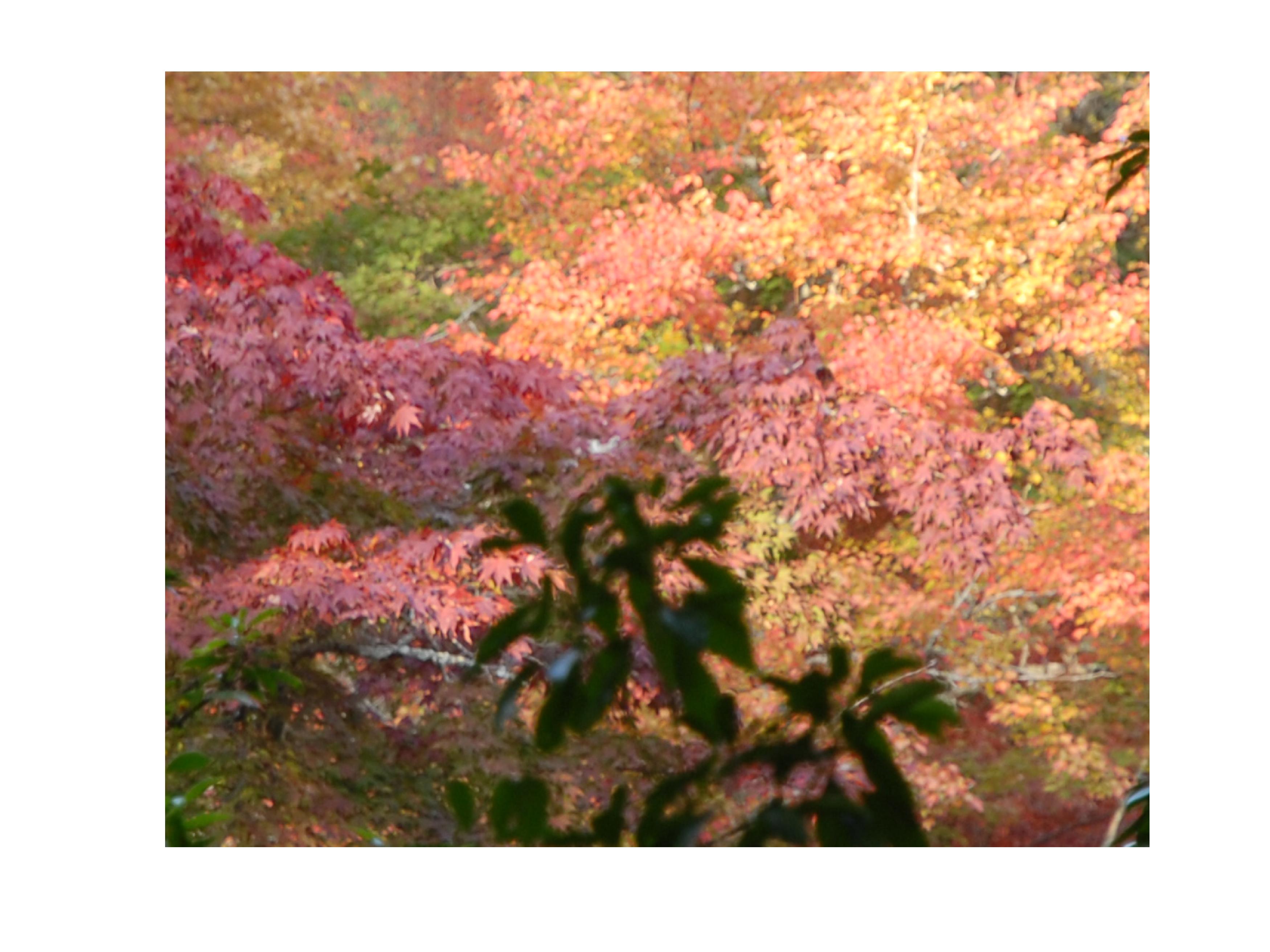}&
		\includegraphics[width=0.18\linewidth]{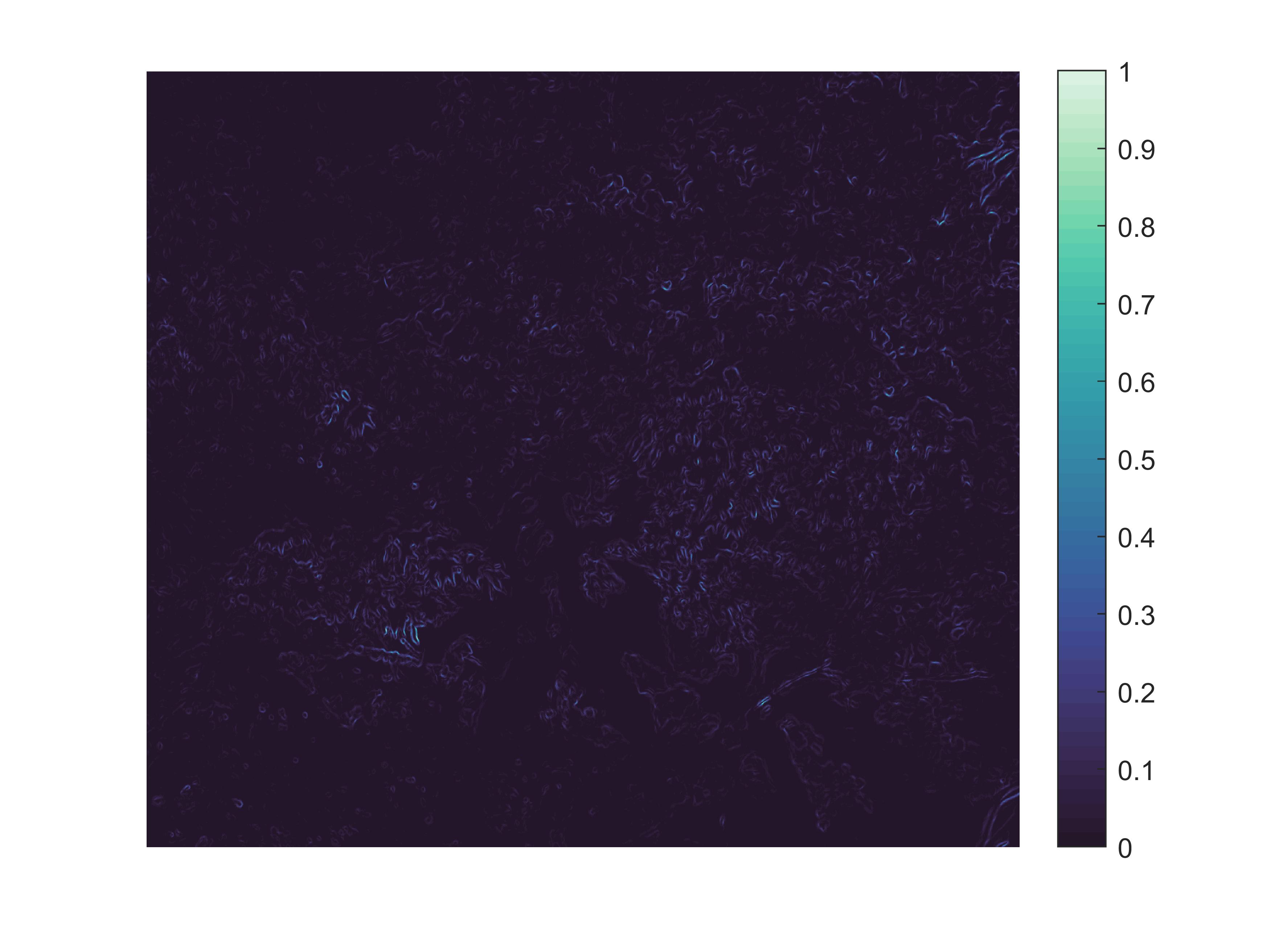}&
		\includegraphics[width=0.18\linewidth]{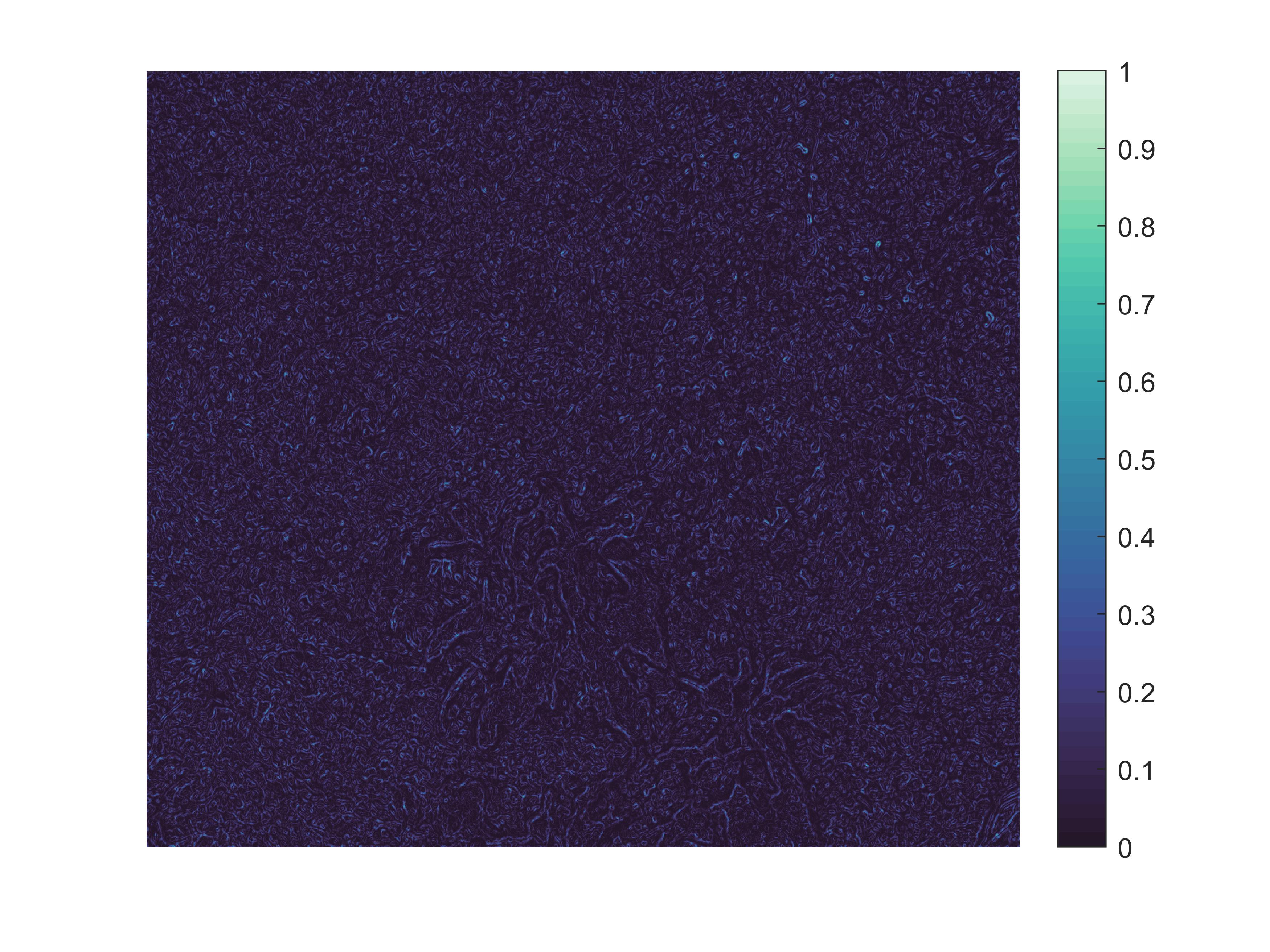}&
		\includegraphics[width=0.18\linewidth]{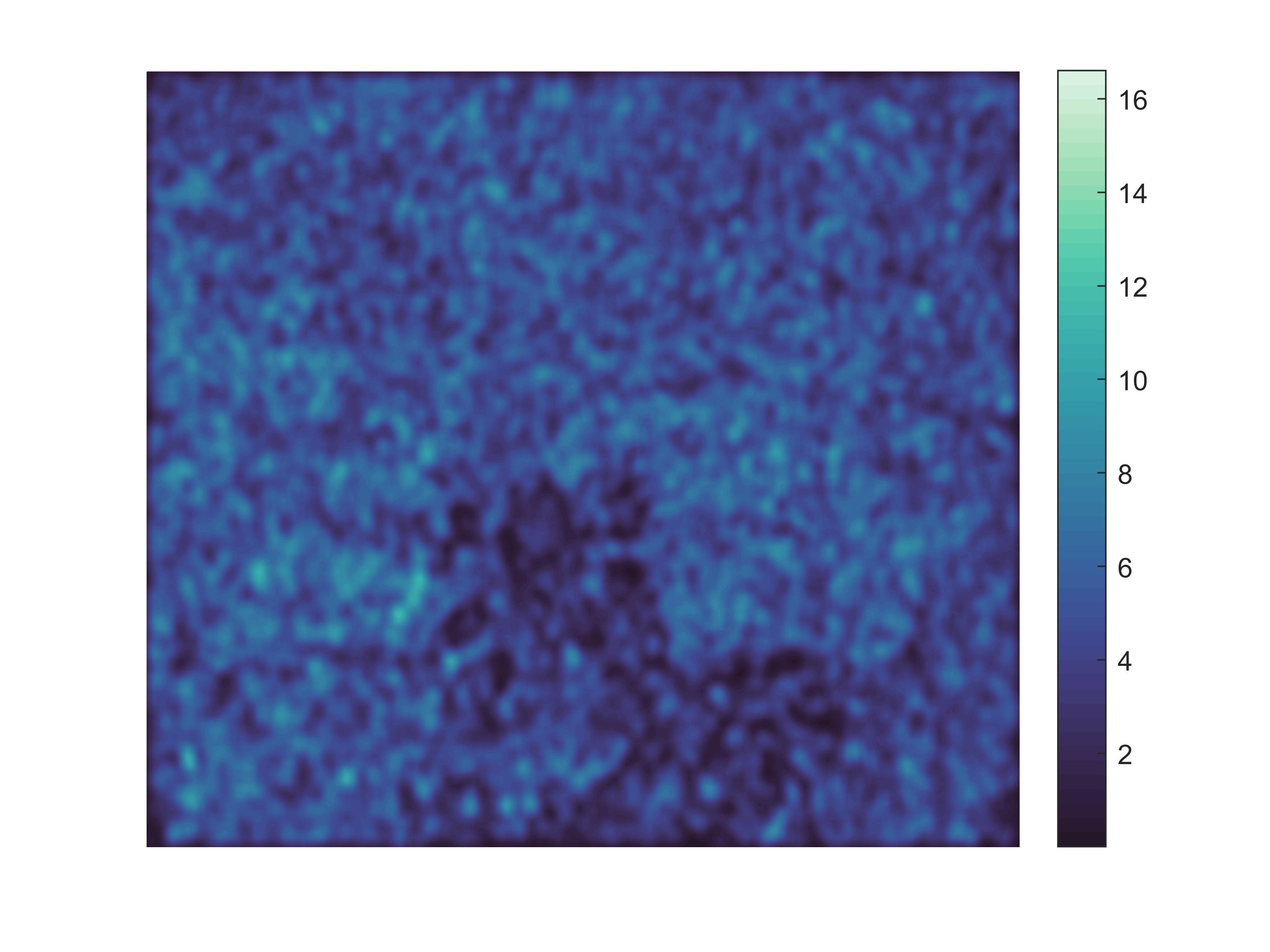}&
		\includegraphics[width=0.18\linewidth]{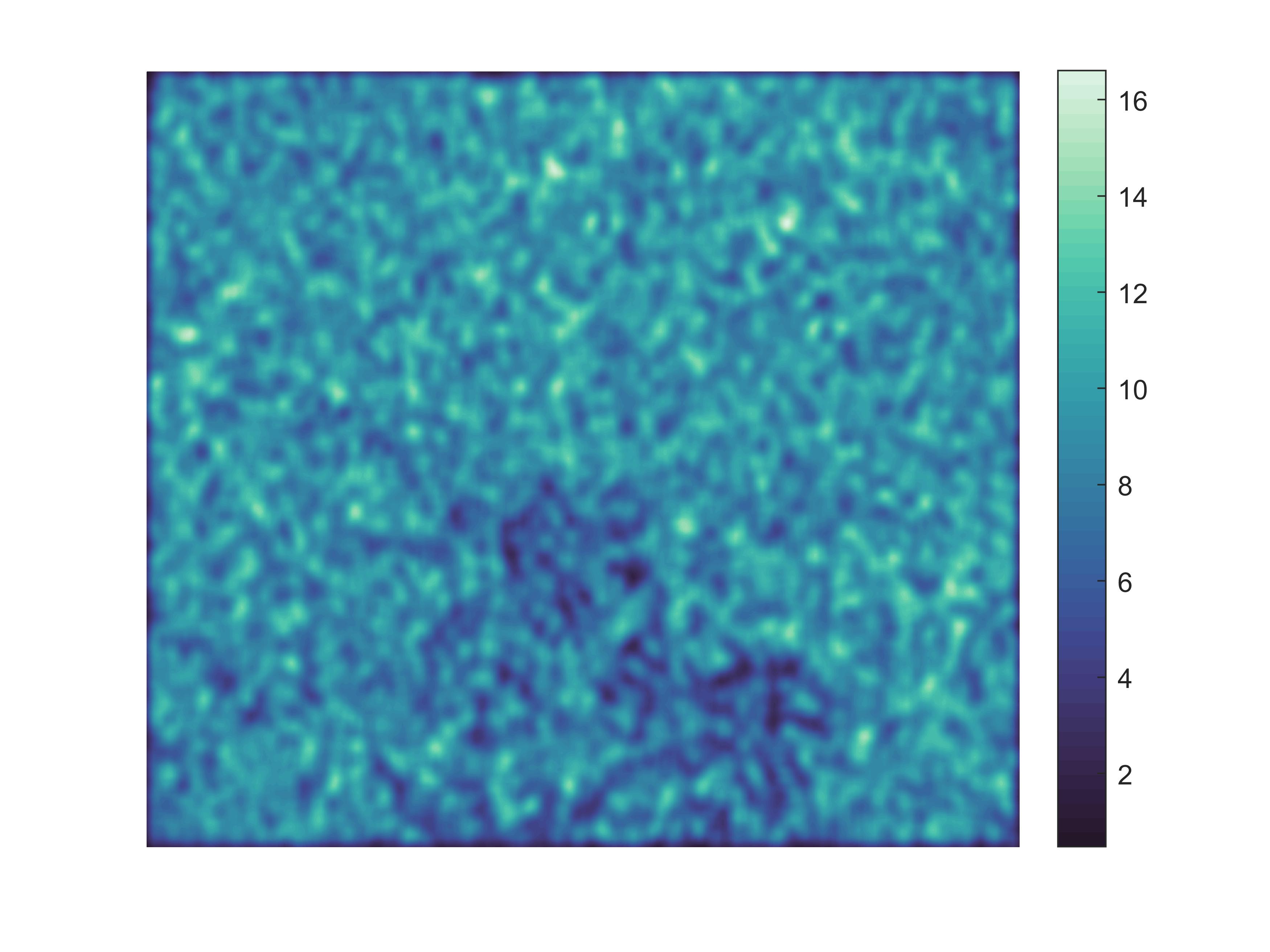}\\
		\includegraphics[width=0.18\linewidth]{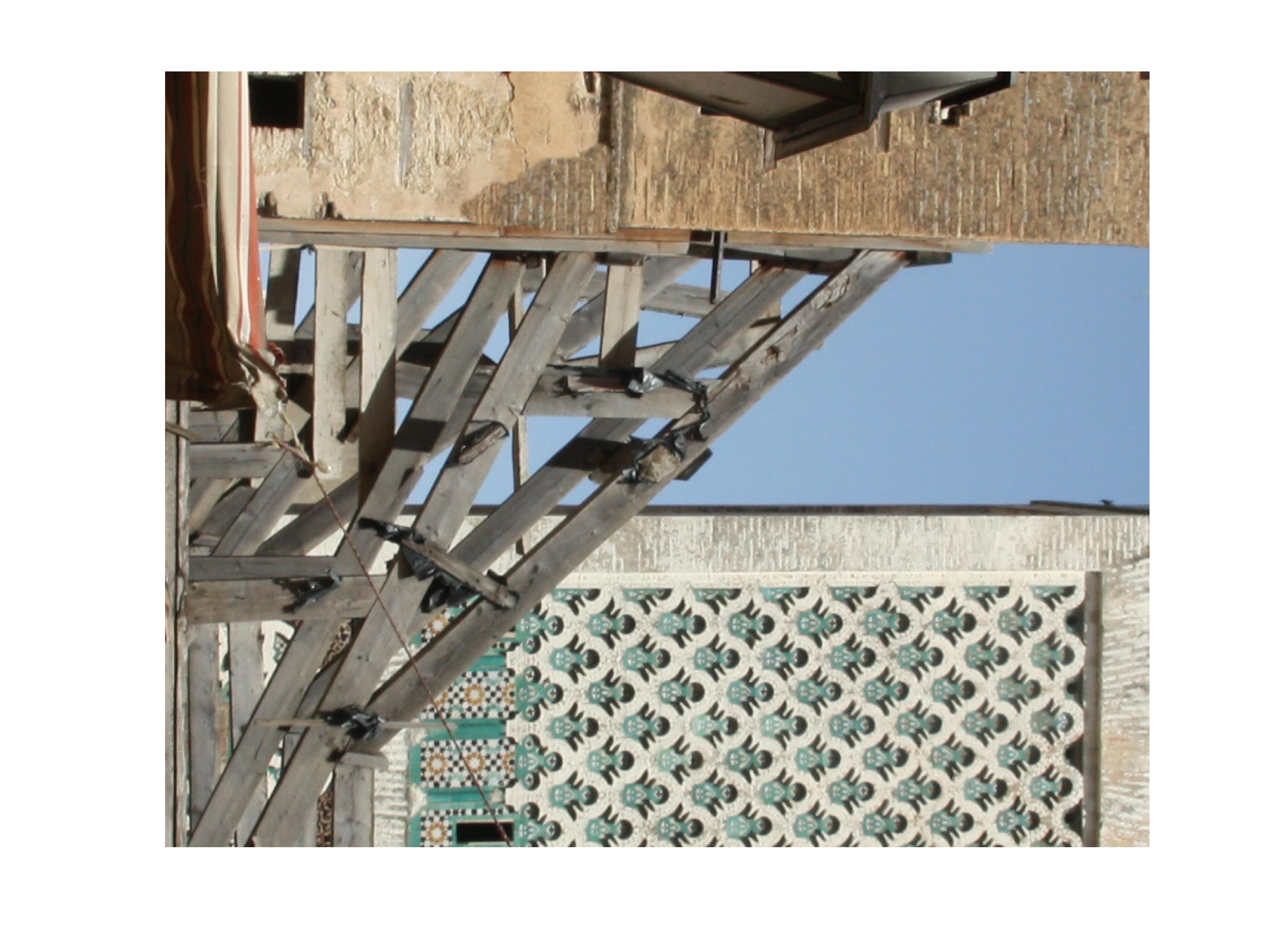}&
		\includegraphics[width=0.18\linewidth]{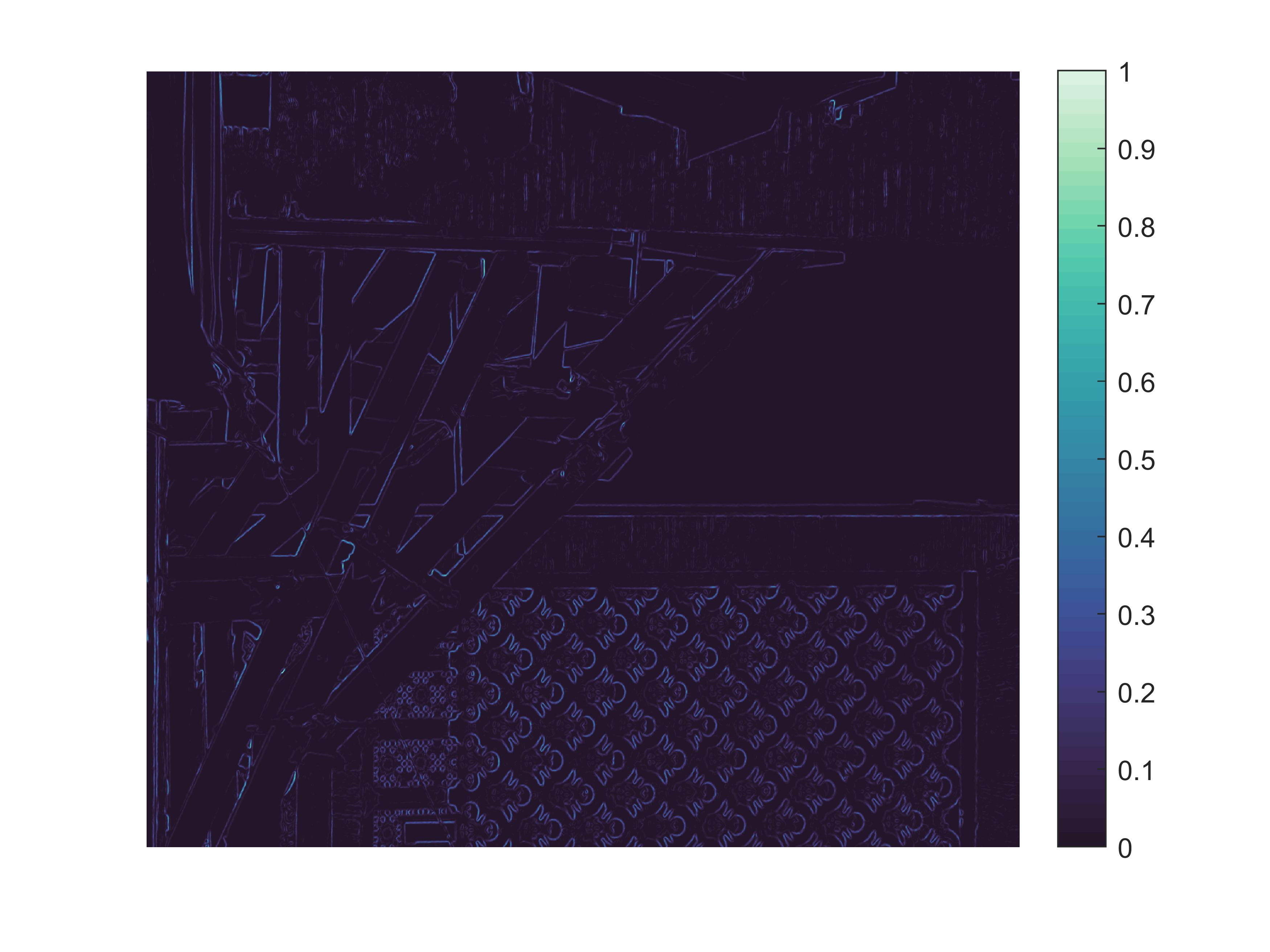}&
		\includegraphics[width=0.18\linewidth]{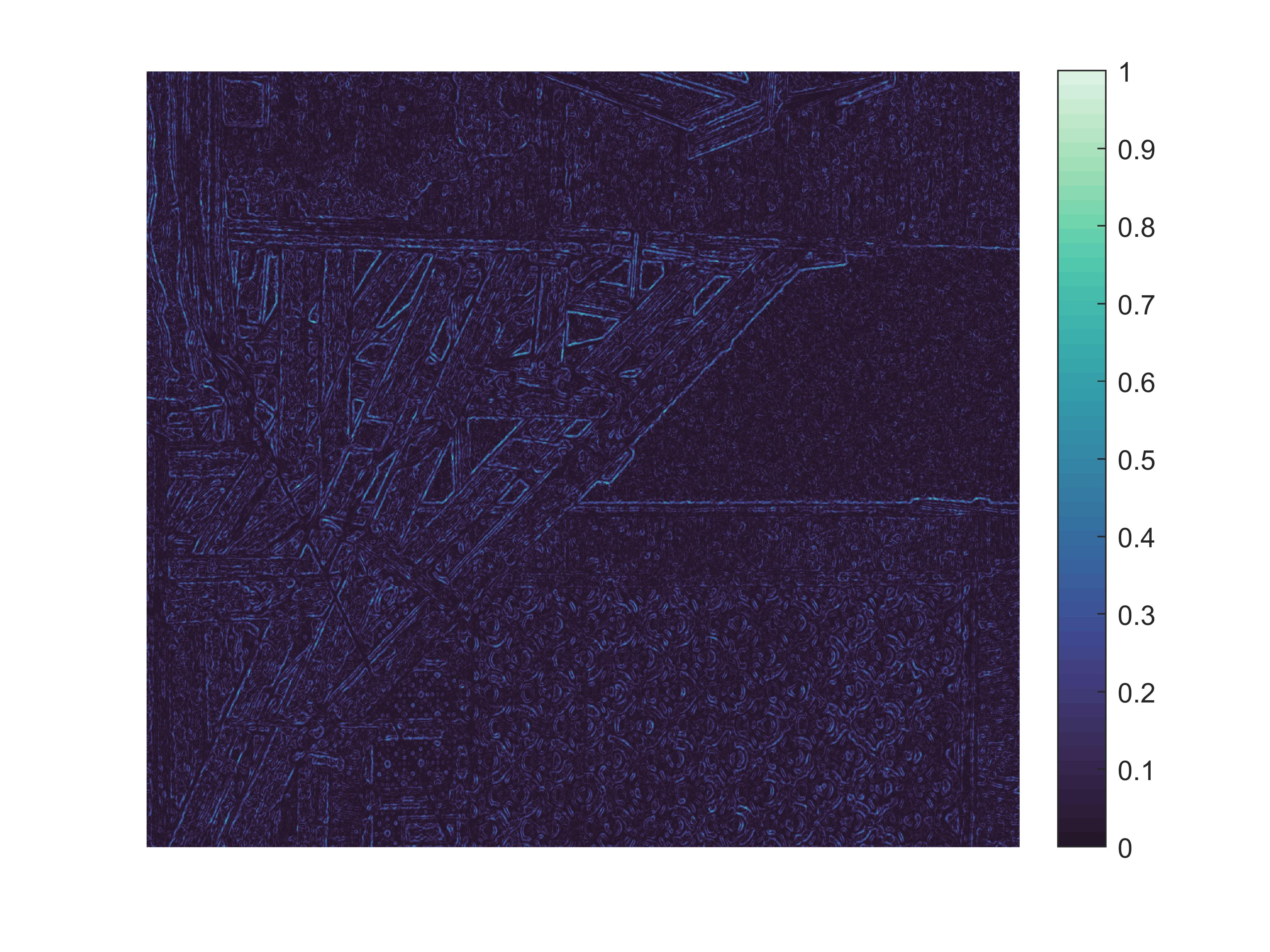}&
		\includegraphics[width=0.18\linewidth]{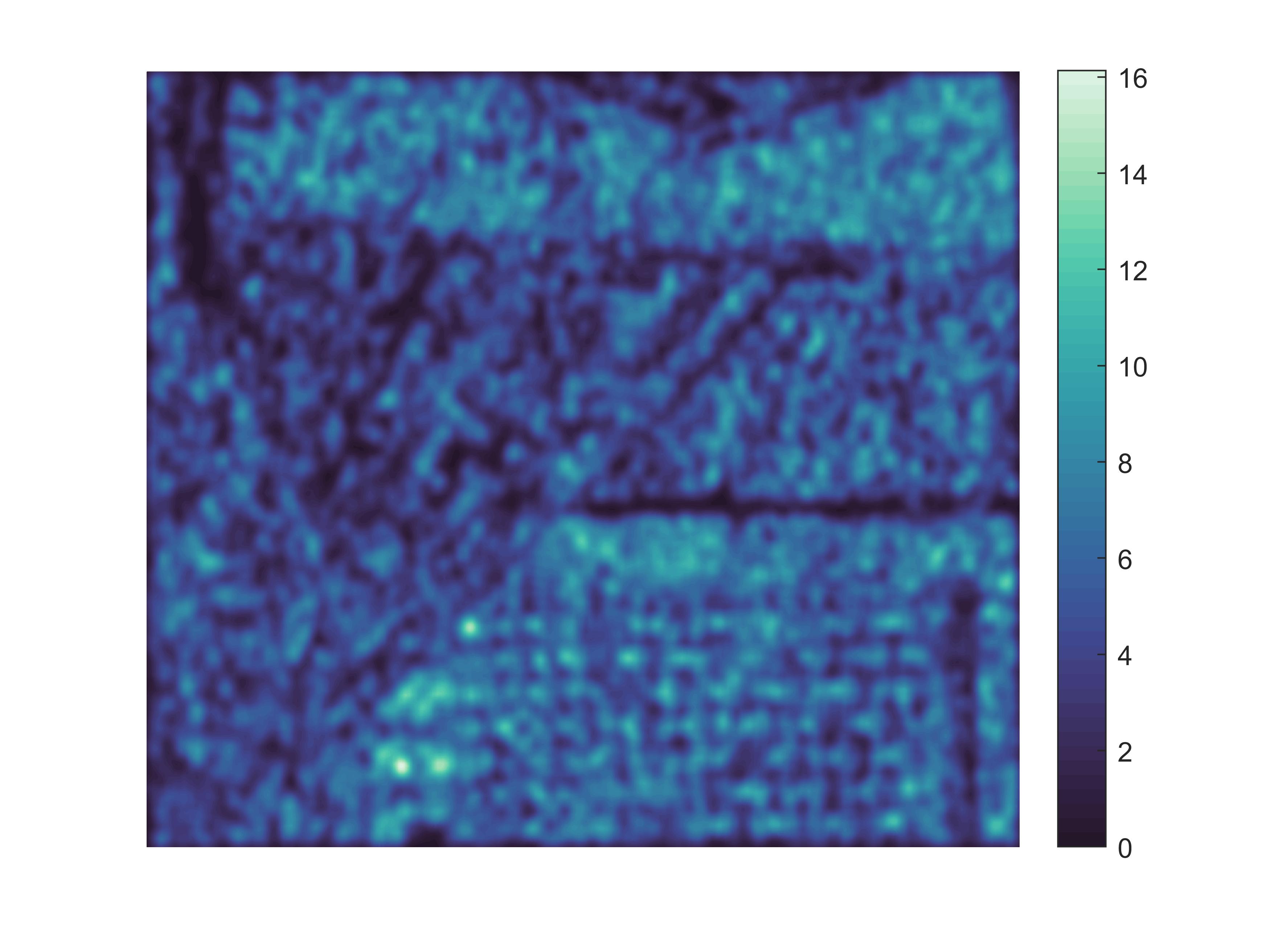}&
		\includegraphics[width=0.18\linewidth]{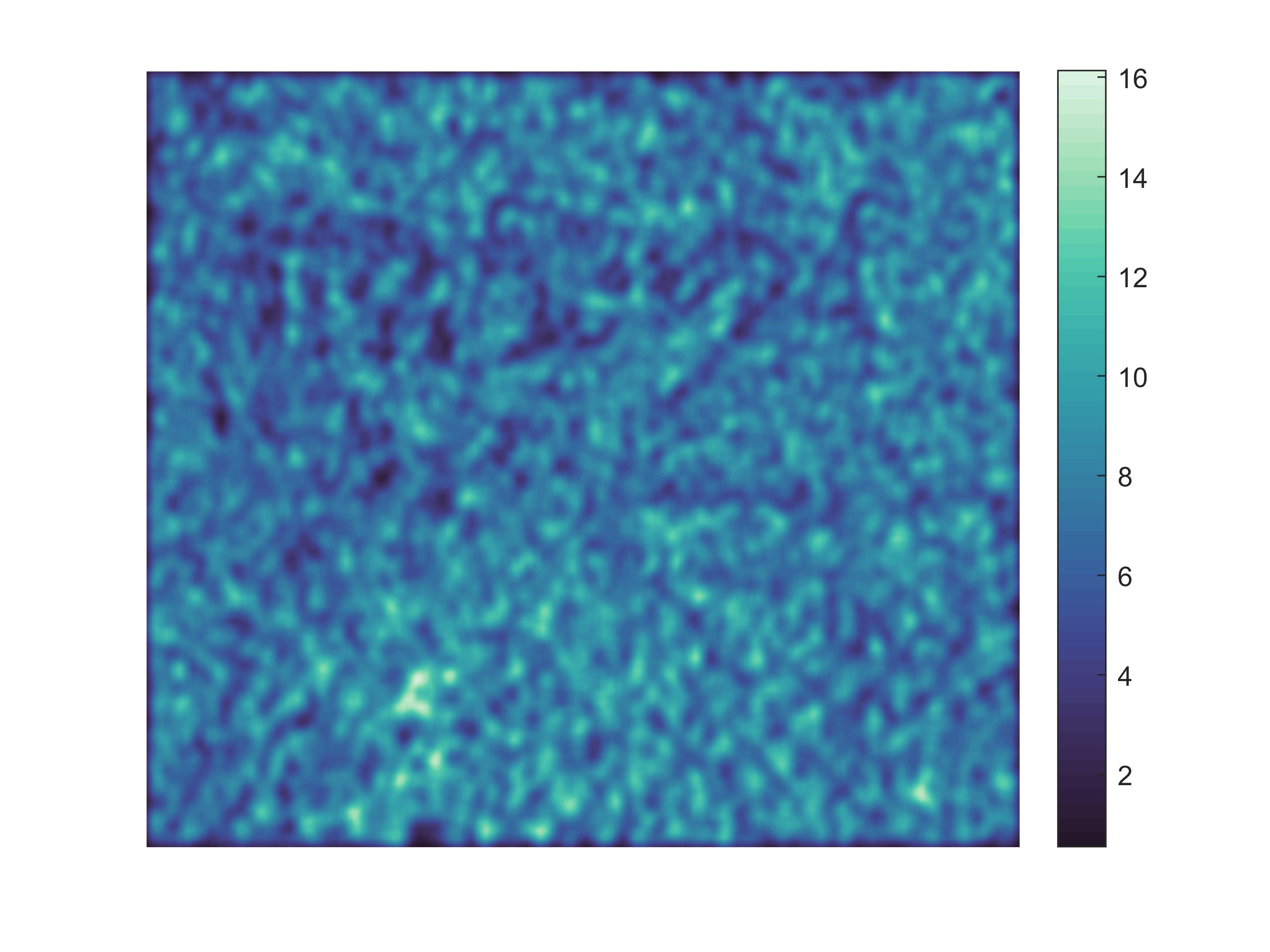}\\
		\includegraphics[width=0.18\linewidth]{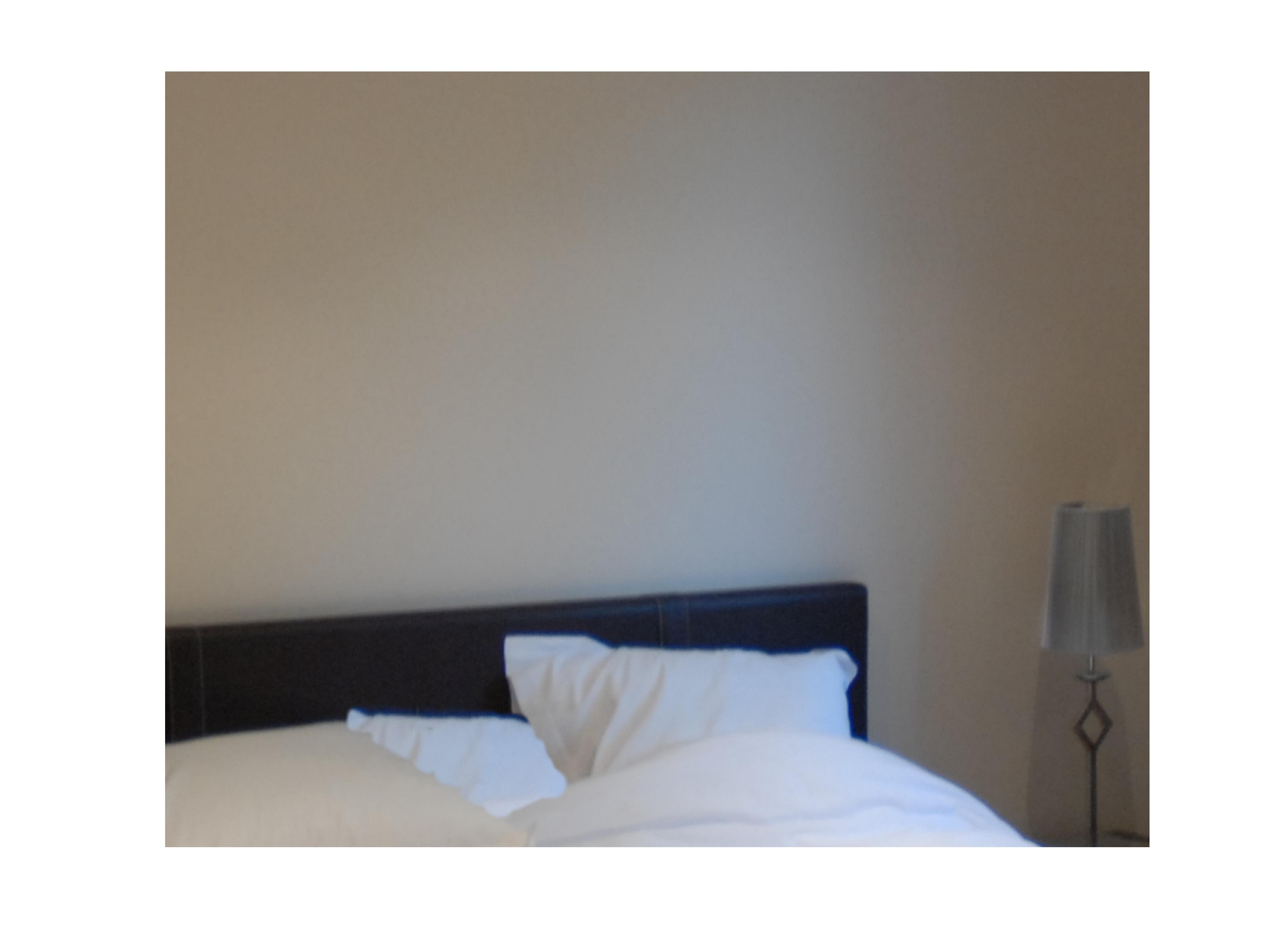}&
		\includegraphics[width=0.18\linewidth]{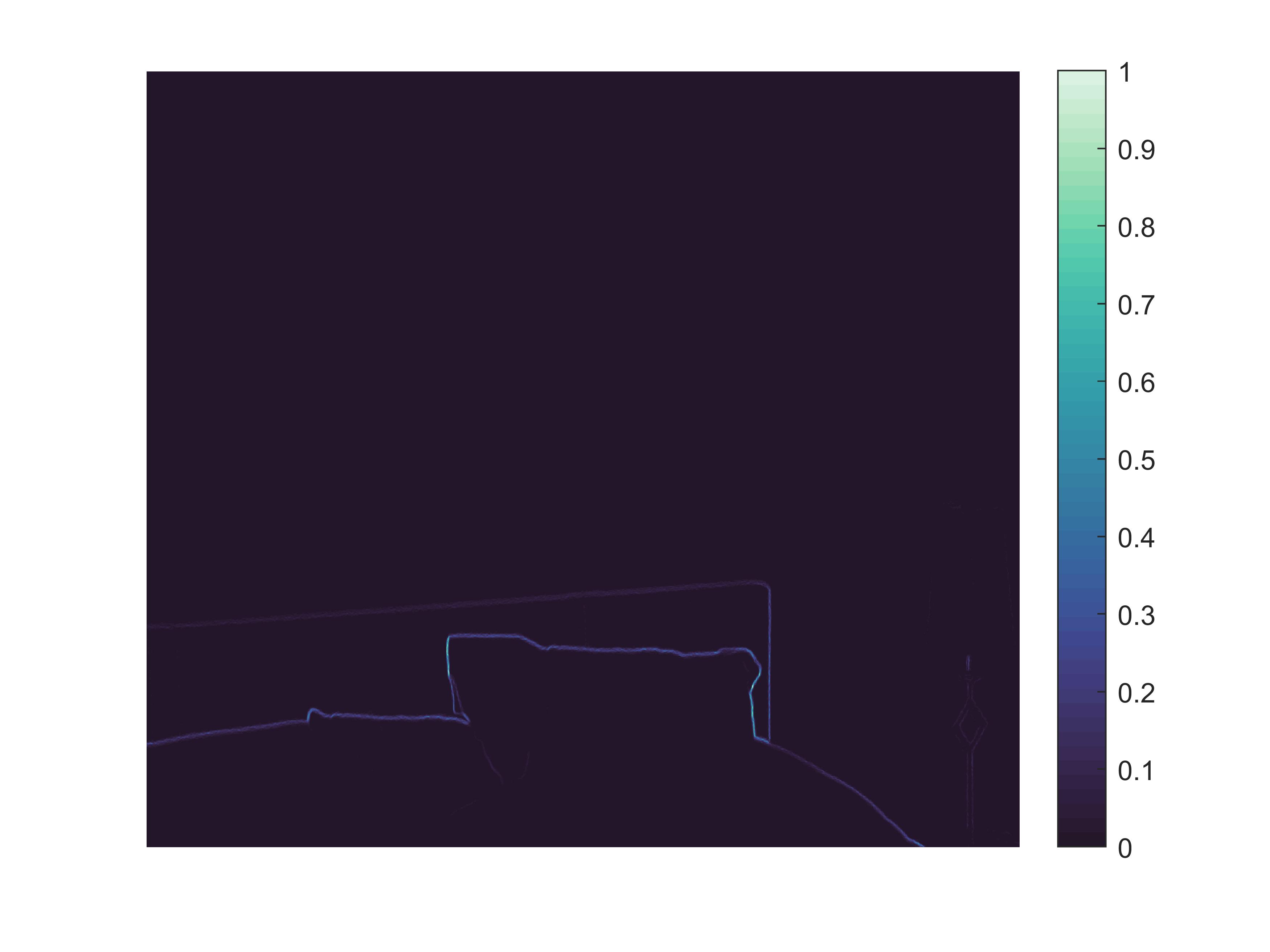}&
		\includegraphics[width=0.18\linewidth]{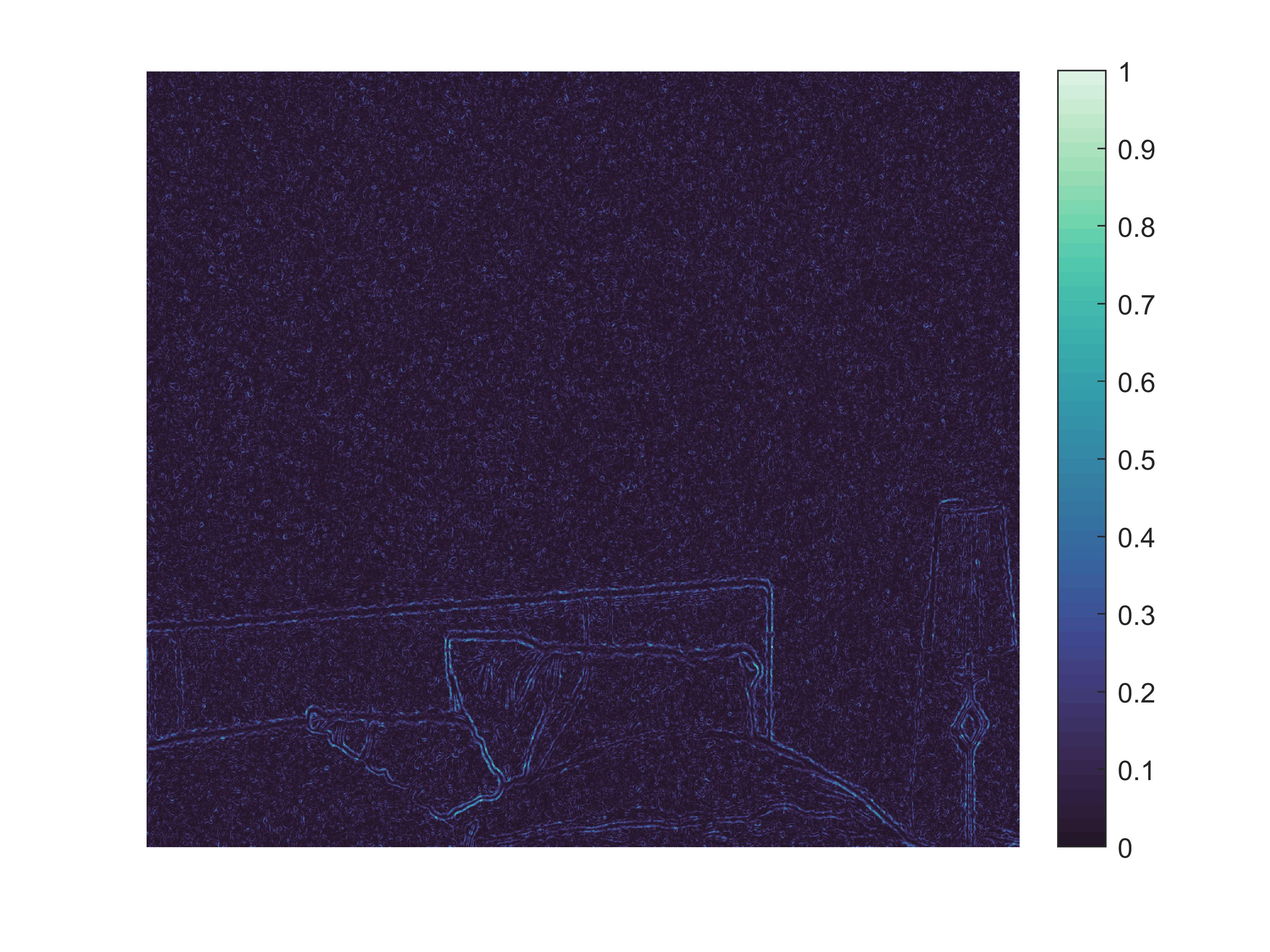}&
		\includegraphics[width=0.18\linewidth]{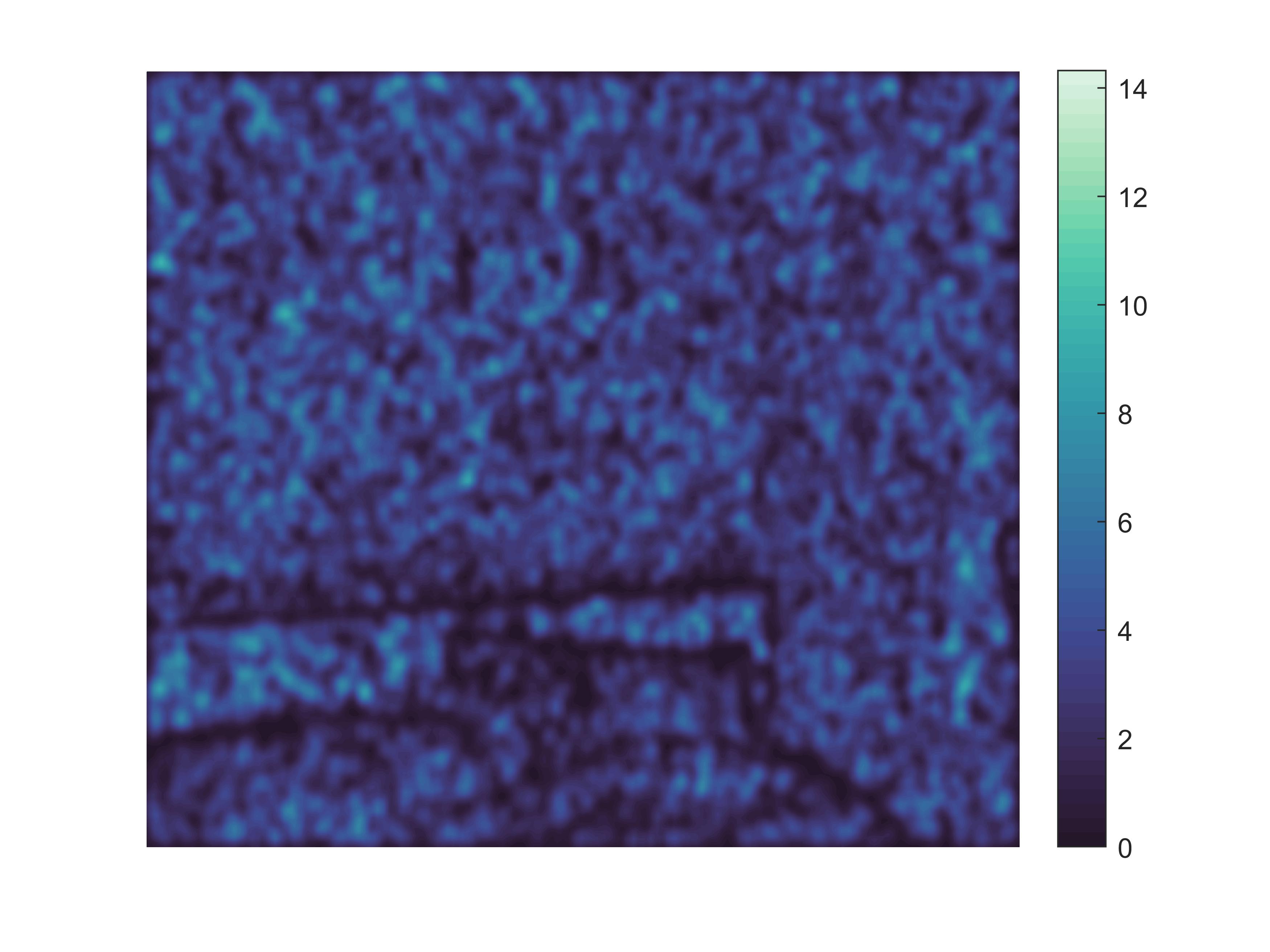}&
		\includegraphics[width=0.18\linewidth]{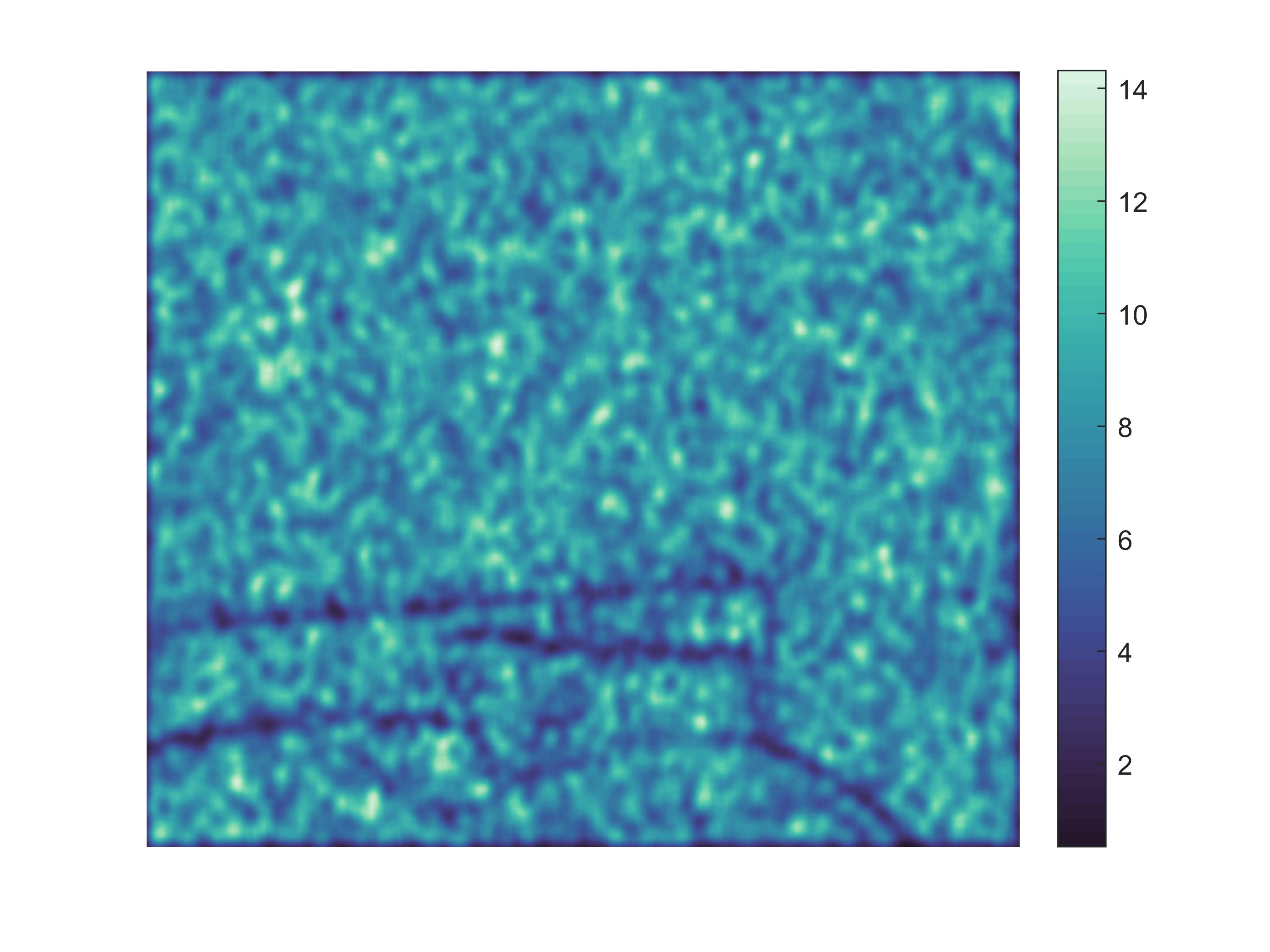}\\
		(a)&(b)&(c)&(d)&(e)
	\end{tabular}
	\caption{Pre-processing results of different types of images. (a) RGB image; (b) the gradient information of gray image; (c) the gradient information of entropy image; (d) the number of keypoints within $ 16 \times 16 $ in gray images; (e) the number of keypoints within $ 16 \times 16 $ in entropy images.}
	\label{Fig2}
\end{figure*}

Compared to gray images, entropy images can more effectively quantify the contrast of local region textures, resulting in a denser distribution of keypoints obtained using entropy images. As shown in Fig. \ref{Fig2} (a), three different types of images are selectd from the GRIP dataset \cite{6}. Then, the gradient information of gray images and entropy images are extracted to represent contrast information. The more high-contrast information, the more keypoints are generated. To enhance the visual experience for readers, the gray image contrast Fig. \ref{Fig2} (b) and entropy image contrast Fig.\ref{Fig2} (c) results are normalized to the range [0, 1] and displayed using pseudo-color images. Obviously, Entropy images have better contrast information.

Besides, there is an overlooked issue at this stage, that is, how do we determine the distribution of keypoints to meet our needs? Some pioneers have done some research. For example, in existing work \cite{1}, the patches with the minimum variance were extracted to ensure that each patch generates 4 keypoints on GRIP \cite{6}. This is necessary because RANSAC estimation \cite{16} requires a minimum of 4 correct matches; As the keypoint-based algorithms just published, Wang et al. \cite{14} normalized the long edge of the low-resolution image to 3000 pixels to obtain a great keypoint distribution.

For the CMFD feature matching task, once the size of the descriptor is larger than the tampering size, there is insufficient evidence to prove that the matching is caused by the similarity of the tampered area. Based on this assumption, this paper believes that the distribution of keypoints should satisfy the requirement to generate 4 keypoints within the feature block size. Since the commonly used block sizes for extracting invariant moment features range from $12\times 12$ to $32\times 32$, this paper adopts $16\times 16$. Fig. \ref{Fig2} (d) and (e) show the number of keypoints in the $ 16 \times 16 $ region of the gray image and entropy image, respectively. Obviously, the entropy image suits this task better since it fulfills the paper's assumptions in more regions. For detailed quantization results, please refer to Fig. \ref{Fig5} (a).

\section{Proposed method}\label{Sec3}
\begin{figure}[htb]
	\centering
	\includegraphics[width=0.9\linewidth]{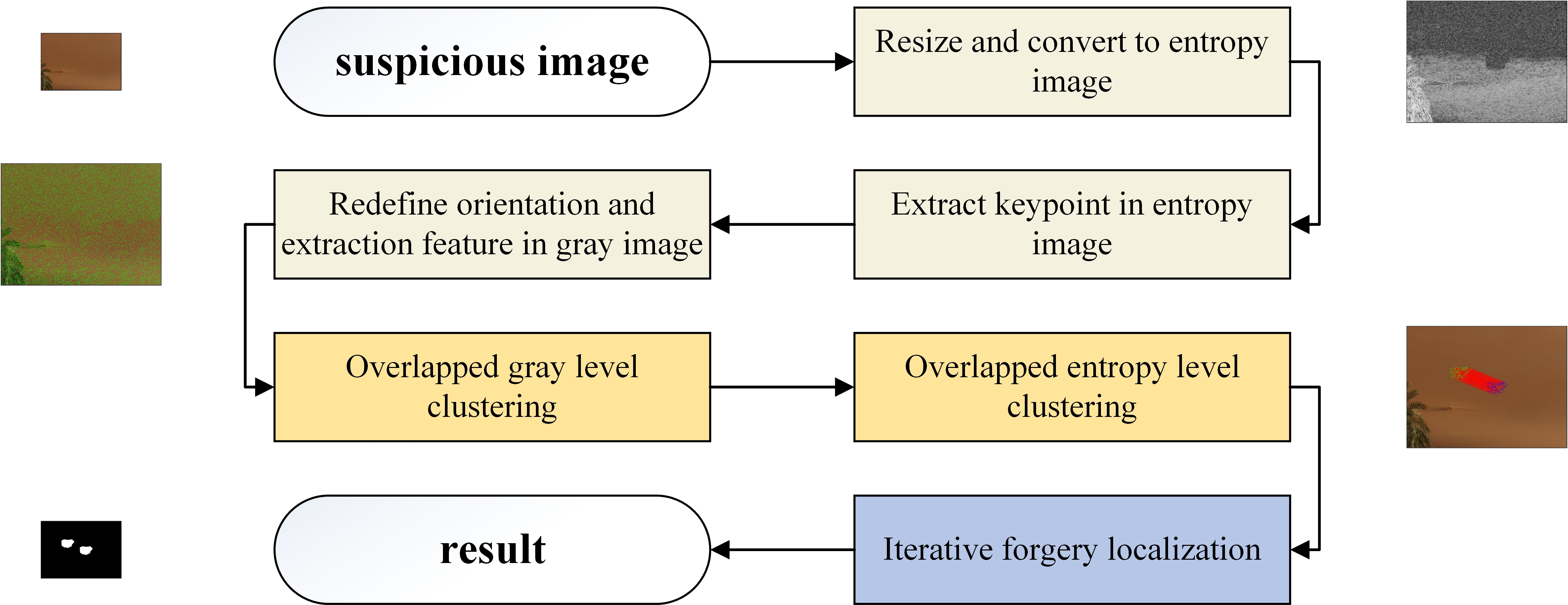}
	\caption{Framework of the proposed algorithm.}
	\label{Fig3}
\end{figure}

As shown in Fig. \ref{Fig3}, the framework of the proposed algorithm consists of three stage. Firstly, the input image goes through the gray box to generate an entropy image, keypoints and corresponding features in the pre-processing stage. Then, keypoints are matched through hierarchical clustering represented by the yellow box in the matching stage. Finally, the method described in literature \cite{1} represented by the blue box will be used in the post-processing stage. This post-processing was designed by fully exploiting the dominant orientation and scale information of each matched keypoint. Experimental results demonstrate that our algorithm achieves a good balance between performance and time efficiency.

\subsection{Pre-processing}\label{sec3_1}

Assuming the input gray image is denoted as $ I_{o} $. Then, the resized  gray image $ I_{gray} $ is defined as:
\begin{equation}
	\label{Eq7}
	{I_{gray}} = {\mathop{\rm imresize}\nolimits} ({I_o},s)
\end{equation}
Here, $ {\mathop{\rm imresize}\nolimits} ( \cdot ) $ and $ s $ denotes image resize function and resize factor, respectively. Bicubic linear interpolation is used in this paper. For instance, if $ I_{o} $ with dimensions $ 1024 \times 728$ and resize it with $ s=2 $, $ I_{gray} $ will have dimensions of $ 2048 \times 1456 $.

Then, the entropy image $ I_E(x, y) $ at position $ (x, y) $ can be expressed as:
\begin{equation}
	\label{Eq8}
	{I_E}(x,y) = \sum\limits_{i = 0}^{255} {{p_i}(x,y) \cdot {{\log }_2}[{p_i}(x,y)]}
\end{equation}
Here, $ p_i(x, y) $ represents the probability of the grayscale value being $ i $ in a square region with radius $ R_E $ around $ (x, y) $. $ I_E $ can be obtained by performing the same steps on $ I_{gray} $.

Next, the SIFT detector is applied to extract keypoints $ \bf{KP_{E}} $ from $ I_E $. $ \bf{KP_{E}} $ can be defined as follows:
\begin{equation}
	\label{Eq9}
	\bf{KP_{E}} = \{ \bf{x_{E}},\bf{y_{E}},\bf{\sigma_E},\bf{\theta_E} \} 
\end{equation}
Here, $ (\bf{x_{E}},\bf{y_{E}}) $ represent the image plane coordinates, $ \bf{\sigma_E} $ denote scale information. The dominant orientation $ \bf{\theta_E} $ of keypoints $ \bf{KP_{E}} $ are then redefined on the gray image through Equation (\ref{Eq6}), and features $ \bf{F} $ are extracted from $ I_{gray} $. Finally, $ \bf{KP} $ can be defined as follows:
\begin{equation}
	\label{Eq10}
	\bf{KP} = \{ \bf{x_{E}},\bf{y_{E}},\bf{\sigma_E},\theta_{gray} \} 
\end{equation}

\subsection{Hierarchical  Keypoint Clustering Matching}\label{Sec3_2}
Gray value is widely used in keypoint clustering algorithms as it effectively represents the fundamental information of an image. The major advantage of these methods is conducted in a much more efficient way without deleting original correct matches. In this paper, an overlapped gray level clustering method \cite{1} is introduced. 

\begin{figure}[htb]
	\centering
	\begin{tabular}{cc}
		\includegraphics[width=0.45\linewidth]{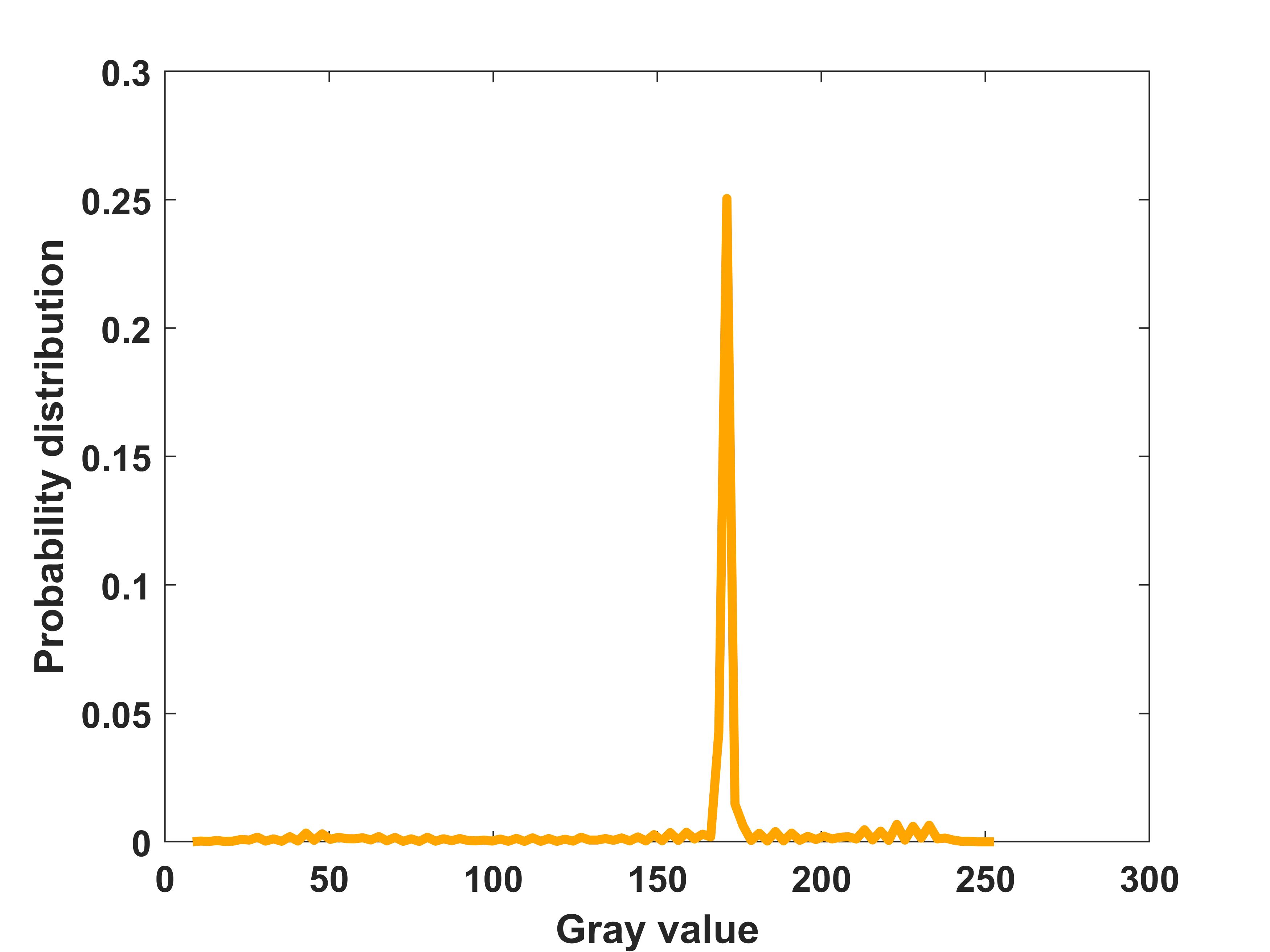} &
		\includegraphics[width=0.45\linewidth]{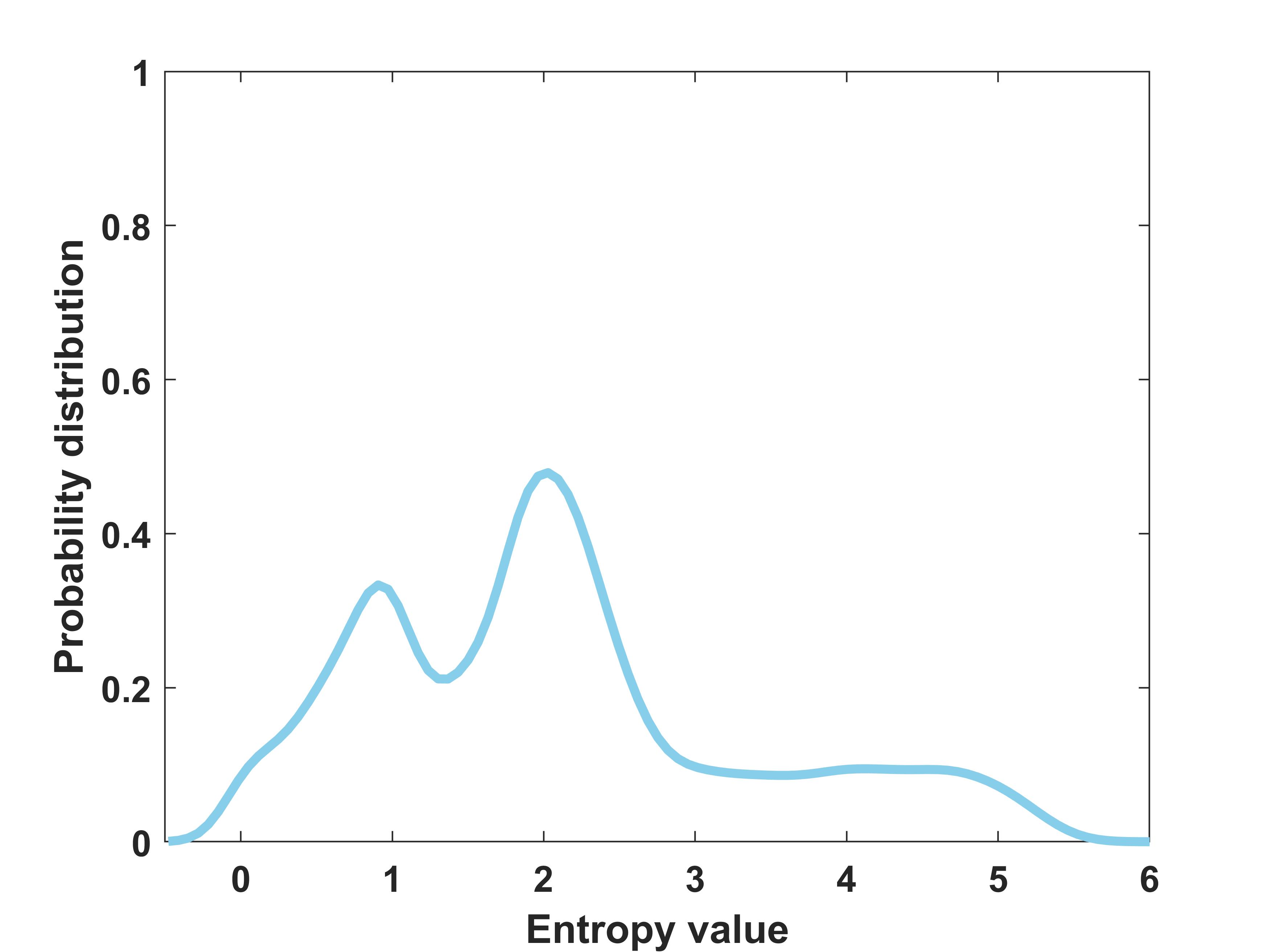} \\
		(a) & (b)
	\end{tabular}
	
	\caption{An example of hierarchical feature point clustering. (a) the gray value distribution of the keypoints; (b) the entropy value distribution of the keypoints.}
	\label{Fig4}
\end{figure}

Although overlapped gray level clustering can be conducted in an effective way, its efficiency may decrease when the gray values of keypoints are concentrated within a specific range. Fig. \ref{Fig4} (a) shows the gray distribution of the keypoints obtained from a certain suspicious image. The gray values of keypoints are concentrated in the range of [150, 200]. Our statistical analysis indicates that around 70\% of keypoints fall within this range. This will greatly reduce the efficiency of clustering matching. 

To overcome this challenge, the overlapped entropy level clustering is developed. From Fig. \ref{Fig4} (b), it is evident that the entropy distribution exhibits a favorable pattern even in cases where the gray distribution is not ideal. Therefore, it is possible to partition the gray values and entropy values of the two-dimensional plane in a reasonable way.

After that, this paper will explain the hierarchical clustering algorithm. Formally, the group of overlapped gray level clustering $ C^u $ can be expressed as:
\begin{equation}
	\label{Eq11}
	\left\{ {\begin{array}{*{20}{l}}
			{ {C^u} = \{ k{p_i}|{I_{gray}}(k{p_i}) \in [a_l^u,a_h^u),k{p_i} \in \bf{KP}\} }\\
			{a_l^u = (u - 1) \cdot (ste{p_1} - ste{p_2})}\\
			{a_h^u = \min (a_l^u + ste{p_1},255)}
	\end{array}} \right.
\end{equation}
Here, $ step_{1} $ represents the interval size, $ step_{2} $ represents the overlapped size ($ step_{1} > step_{2} $). The number of gray level groups is denoted as $ N_{u} $, and can be computed using the following equation:
\begin{equation}
	\label{Eq12}
	N_{u} = \left\lceil {\frac{{255 - step_{1}}}{{step_{1} - step_{2}}}} \right\rceil  + 1
\end{equation}

The overlapped entropy level clustering algorithm can be expressed as follows:
\begin{equation}
	\label{Eq13}
	\left\{ {\begin{array}{*{20}{l}}
			{{C^{u,v}} = \left\{ {k{p_i}|{I_E}(k{p_i}) \in [a_l^{u,v},a_h^{u,v}],k{p_i} \in {C^u}} \right\}}\\
			{a_l^{u,v} = \max (0,(v - 1) \cdot ste{p_3} - ste{p_4})}\\
			{a_h^{u,v} = \min (7,v \cdot ste{p_3} + ste{p_4})}
	\end{array}} \right.
\end{equation}

Due to the fact that entropy values are controlled by $ R_E $, it is evident that all entropy values are distributed within the range of [0, 7](In this paper, $ R_E=3 $, so the theoretical maximum value of entropy is $ - {\log _2}(P) =  - {\log _2}(1/{(2 \cdot {R_E} + 1)^2}) \approx 5.61 $). The number of entropy level groups for group $ C^u $ is denoted as $ N_{u,v} $. Obviously, $ N_{u,v} $ can be computed using the following equation:
\begin{equation}
	\label{Eq14}
	{N_{u,v}} = \left\lceil {\frac{{7 - ste{p_4}}}{{ste{p_3}}}} \right\rceil
\end{equation}

Once hierarchical keypoint clustering is implemented, the next step involves matching $ C^{u,v} $, while ensuring that the matching process meets the following condition:
\begin{equation}
	\label{Eq15}
	{d_i} < T \cdot {d_{i + 1}},i = 2, \cdots {n_{u,v}}
\end{equation}
Here, $ n_{u,v} $ represents the number of keypoints within $ C^{u,v} $'s group, $ d $ represents the ascending distance between a certain feature and all features within $ C^{u,v} $. Obviously, $ d_1 $ represents the distance between the feature and itself, so $ d_1=0 $. For the matching stage, this paper maintains Amerini's \cite{15} setting ($ T=0.5 $). For the post-processing stage, we follow the iterative forgery localization algorithm used in reference \cite{1}.

\section{Experiments}\label{Sec4}
In this section, our proposed method is evaluated through a series of simulation experiments. All the experiments are done using MATLAB R2018a under Microsoft Windows. The PC used for testing has 2.30 GHz CPU and 16 GB RAM. Our code is available at https://github.com/LUZW1998/CMFDUEI.
\subsection{Datasets}\label{Sec4_1}
This paper validates the proposed algorithm using two public datasets, with details as follows:
\begin{itemize}
	\item GRIP: This dataset \cite{6} contains 80 tampered images and 80 original images, all of which are in the size of $ 1024 \times 768 $ pixels.
	\item CMH: This dataset \cite{2} consists of 108 tampered images with resolutions ranging from $ 845 \times 634 $ to $ 3888 \times 2592 $ pixels.
\end{itemize}

For details on the tampering included in each dataset, please see Table \ref{Tab1}.
\begin{table}[htb]
	\centering
	\caption{Tampering types of dataset}
	\label{Tab1}
	\begin{tabular}{|c|c|c|c|} \hline 
		dataset	&	translation	&	Rotation	&	Scaling\\ \hline
		GRIP		&	\textbf{\checkmark}	&				&	\\ \hline
		CMH	&	\textbf{\checkmark}	&	\textbf{\checkmark}	&	\textbf{\checkmark}\\ \hline
	\end{tabular}
\end{table}

\begin{figure*}[h]
	\centering
	\begin{tabular}{cccc}
		\includegraphics[width=0.23\linewidth]{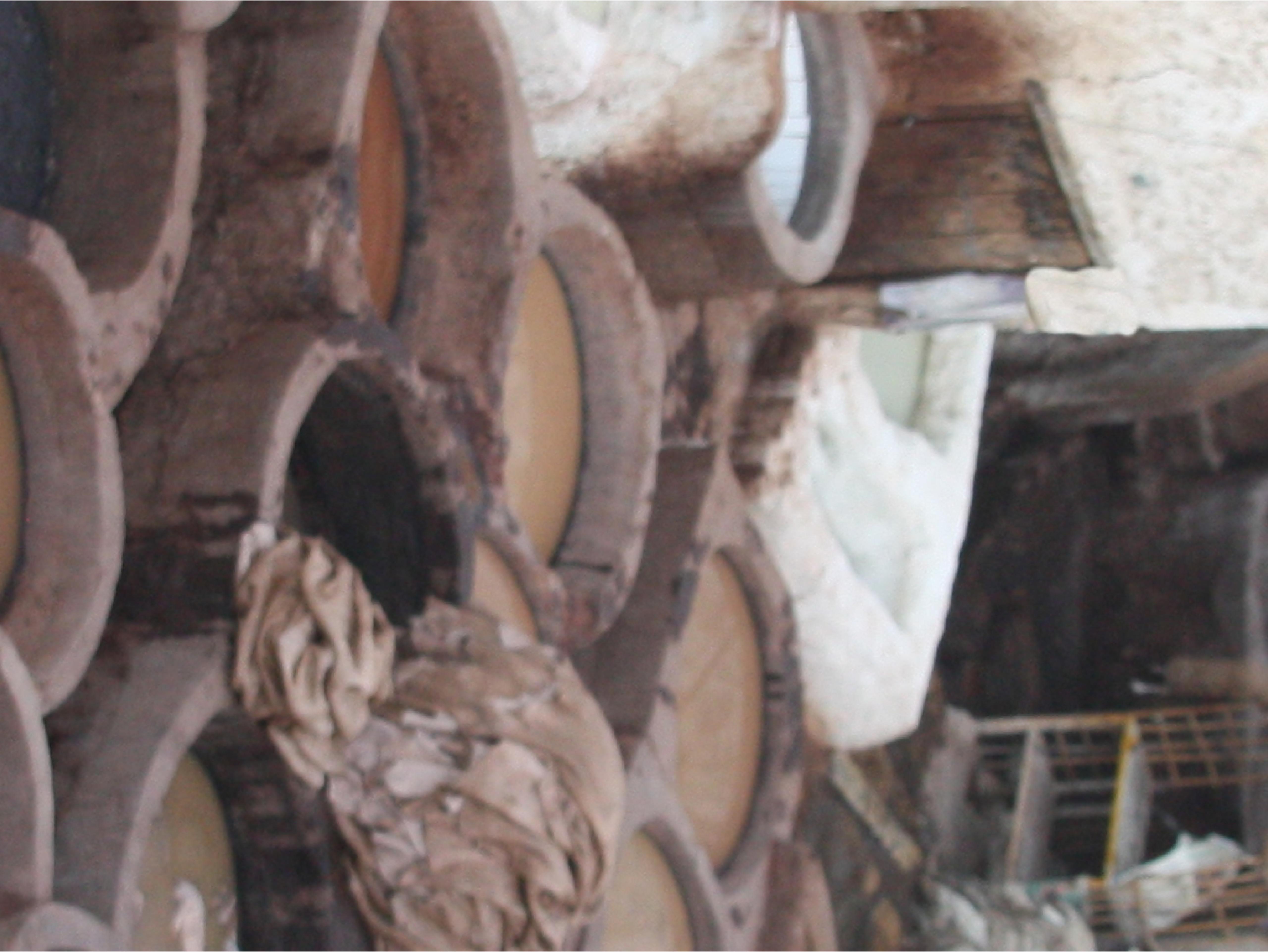} &
		\includegraphics[width=0.23\linewidth]{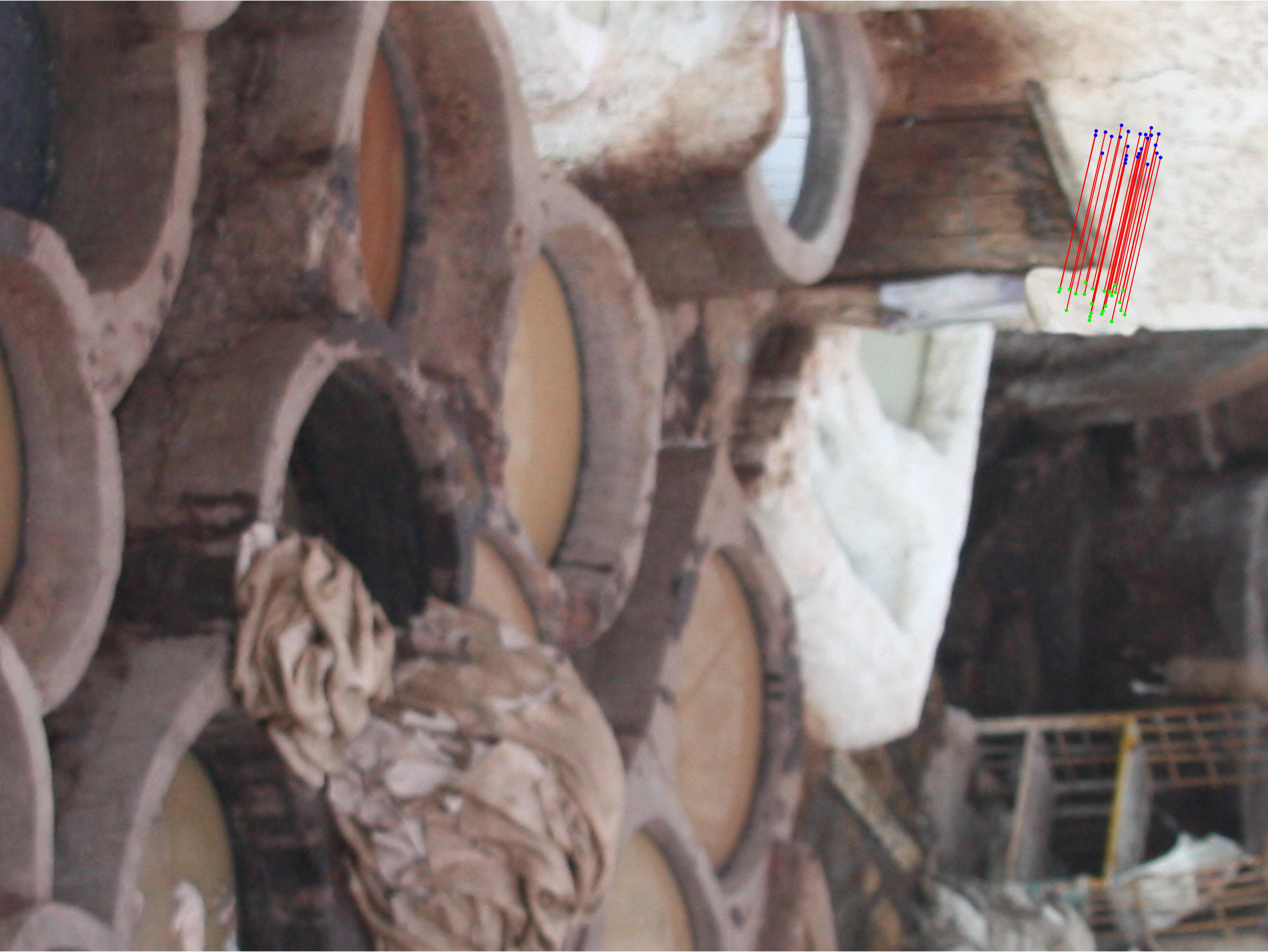} &
		\includegraphics[width=0.23\linewidth]{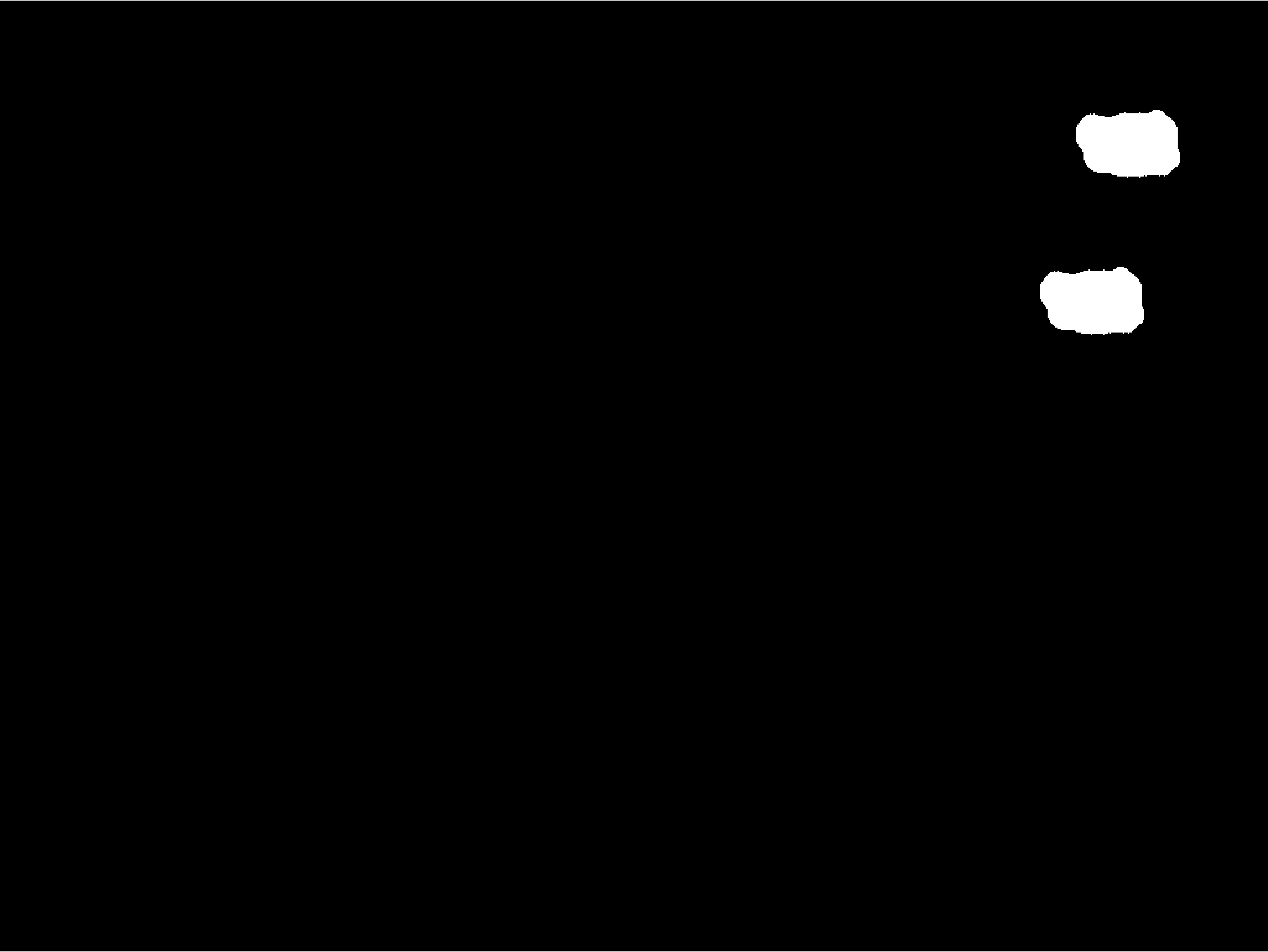} &
		\includegraphics[width=0.23\linewidth]{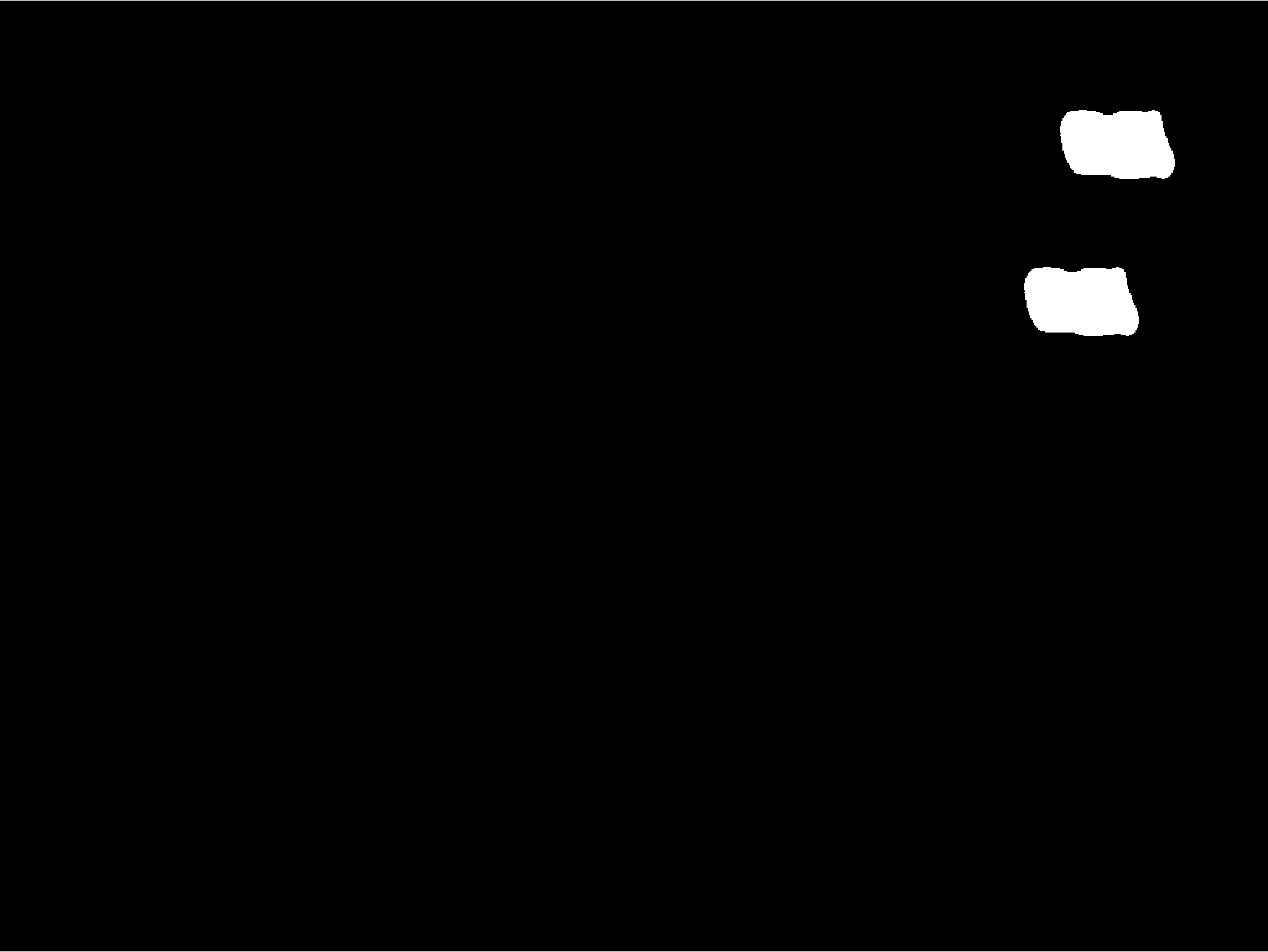} \\
		\includegraphics[width=0.23\linewidth]{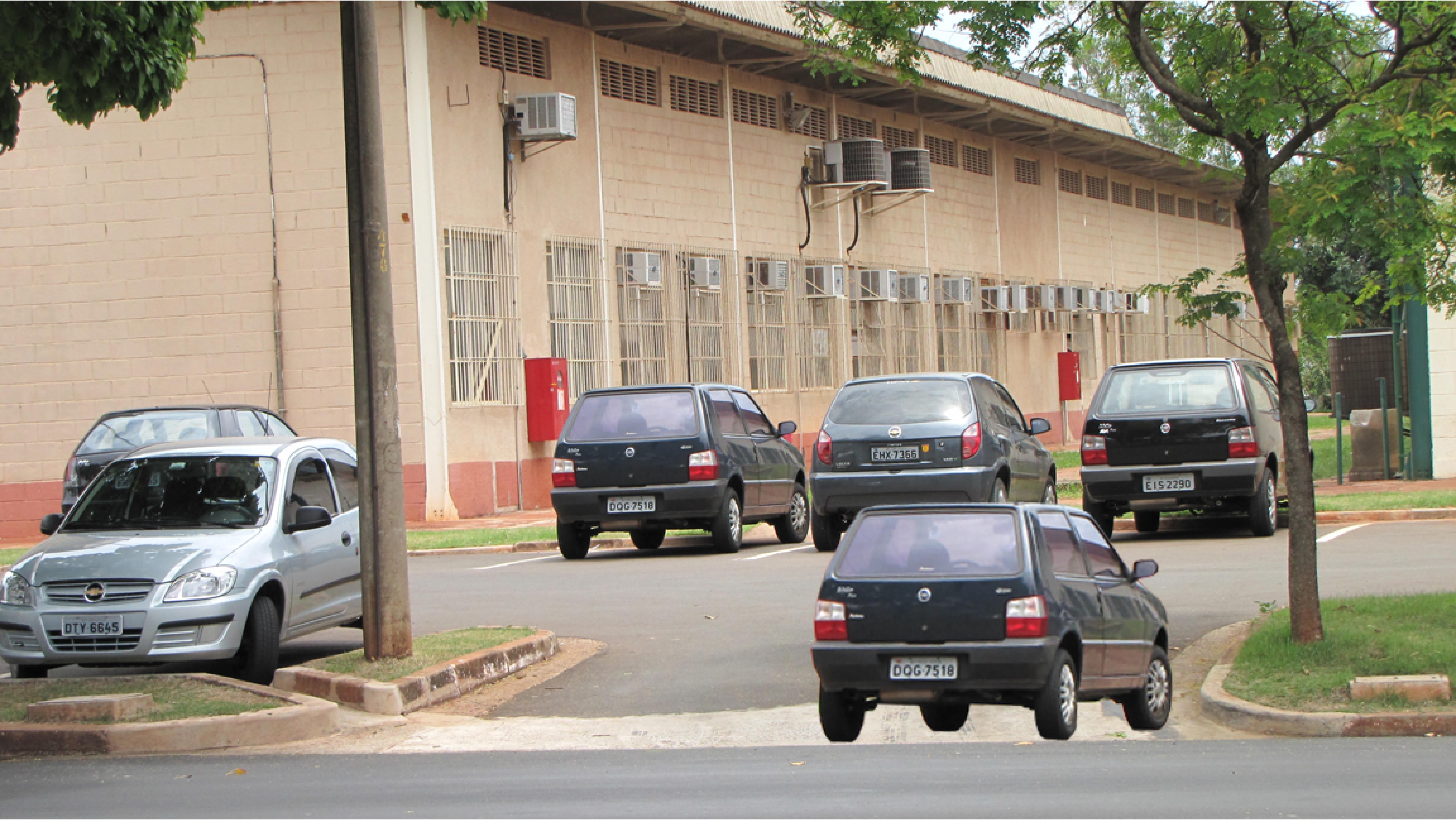} &
		\includegraphics[width=0.23\linewidth]{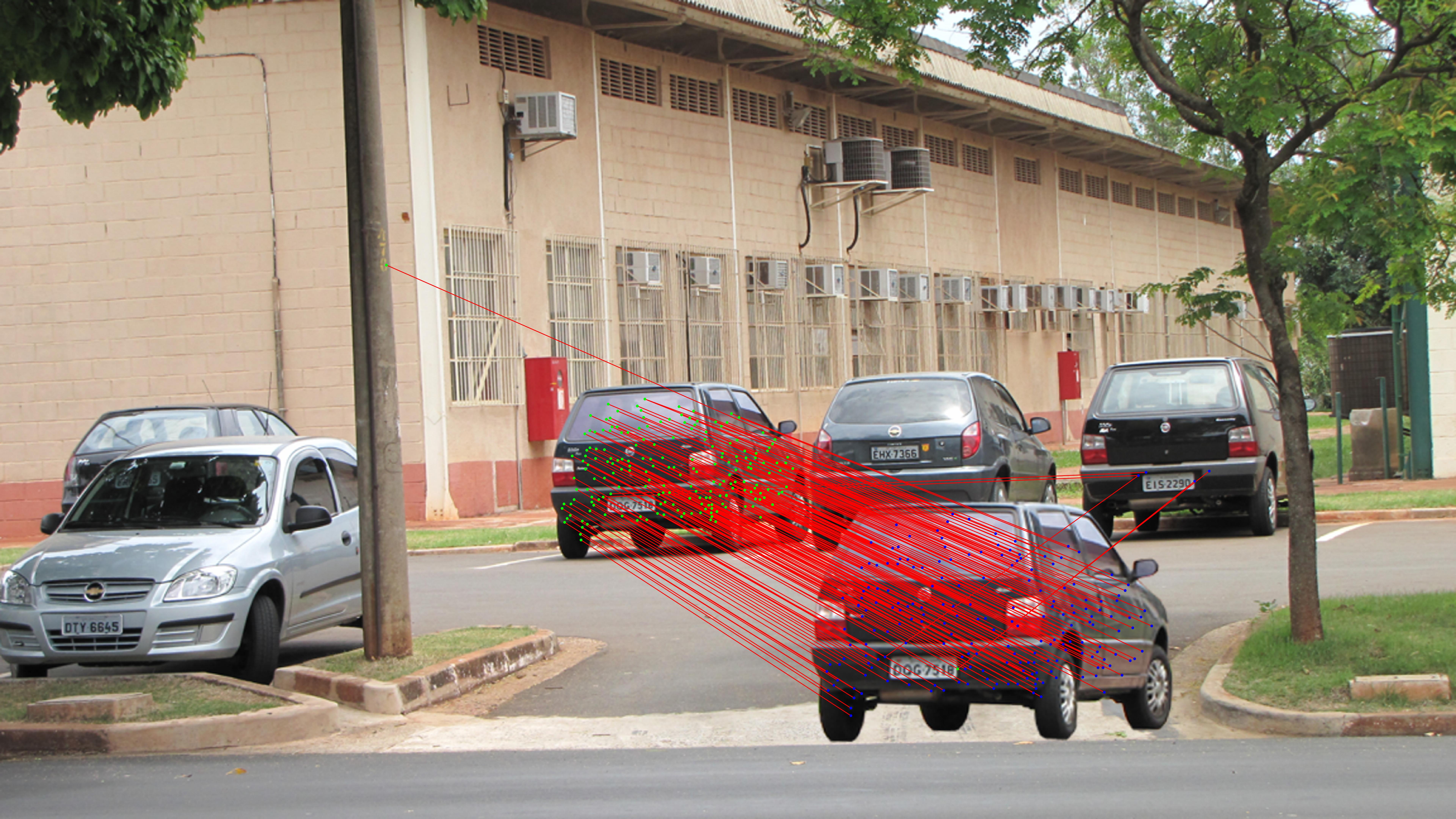} &
		\includegraphics[width=0.23\linewidth]{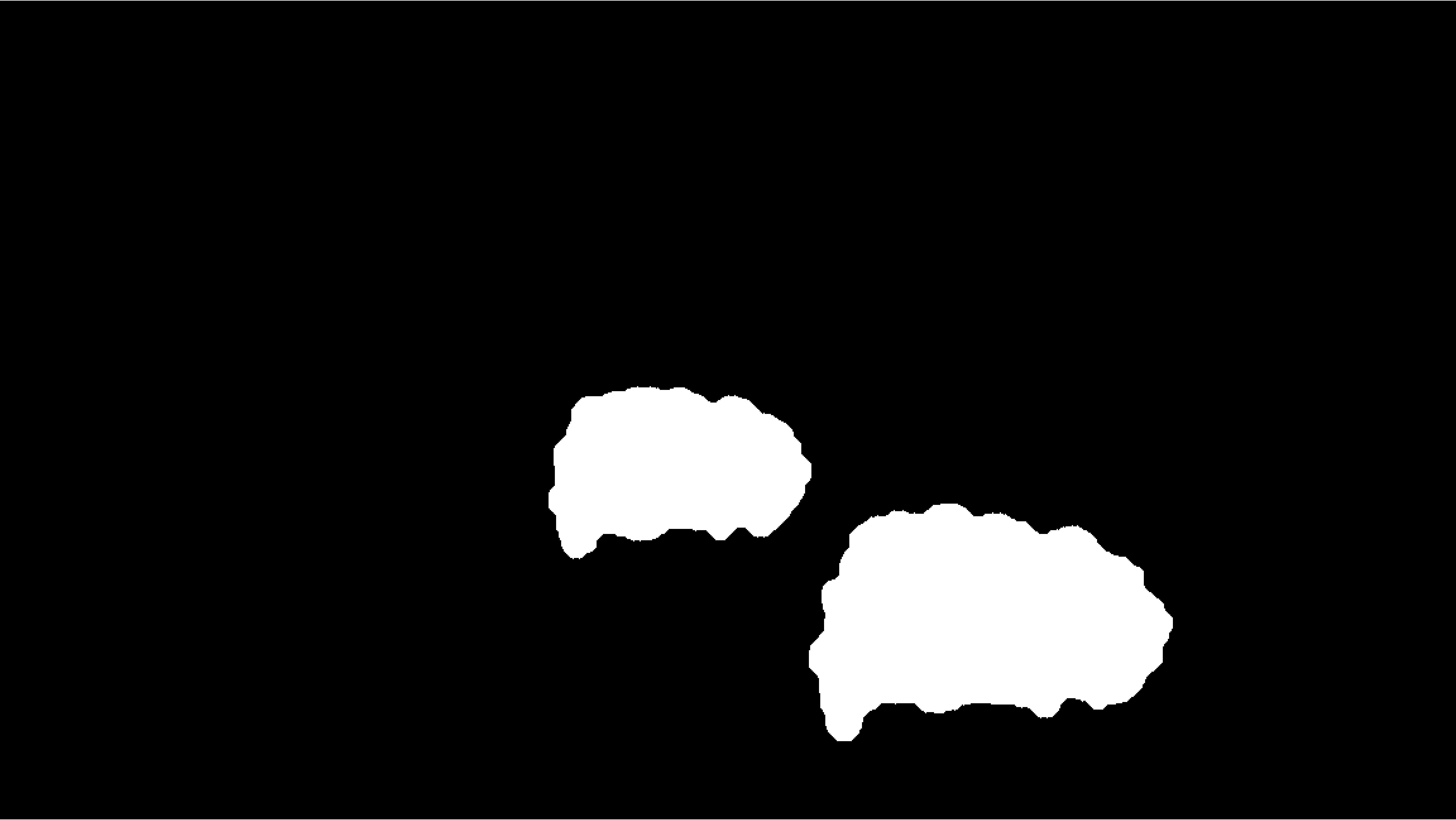} &
		\includegraphics[width=0.23\linewidth]{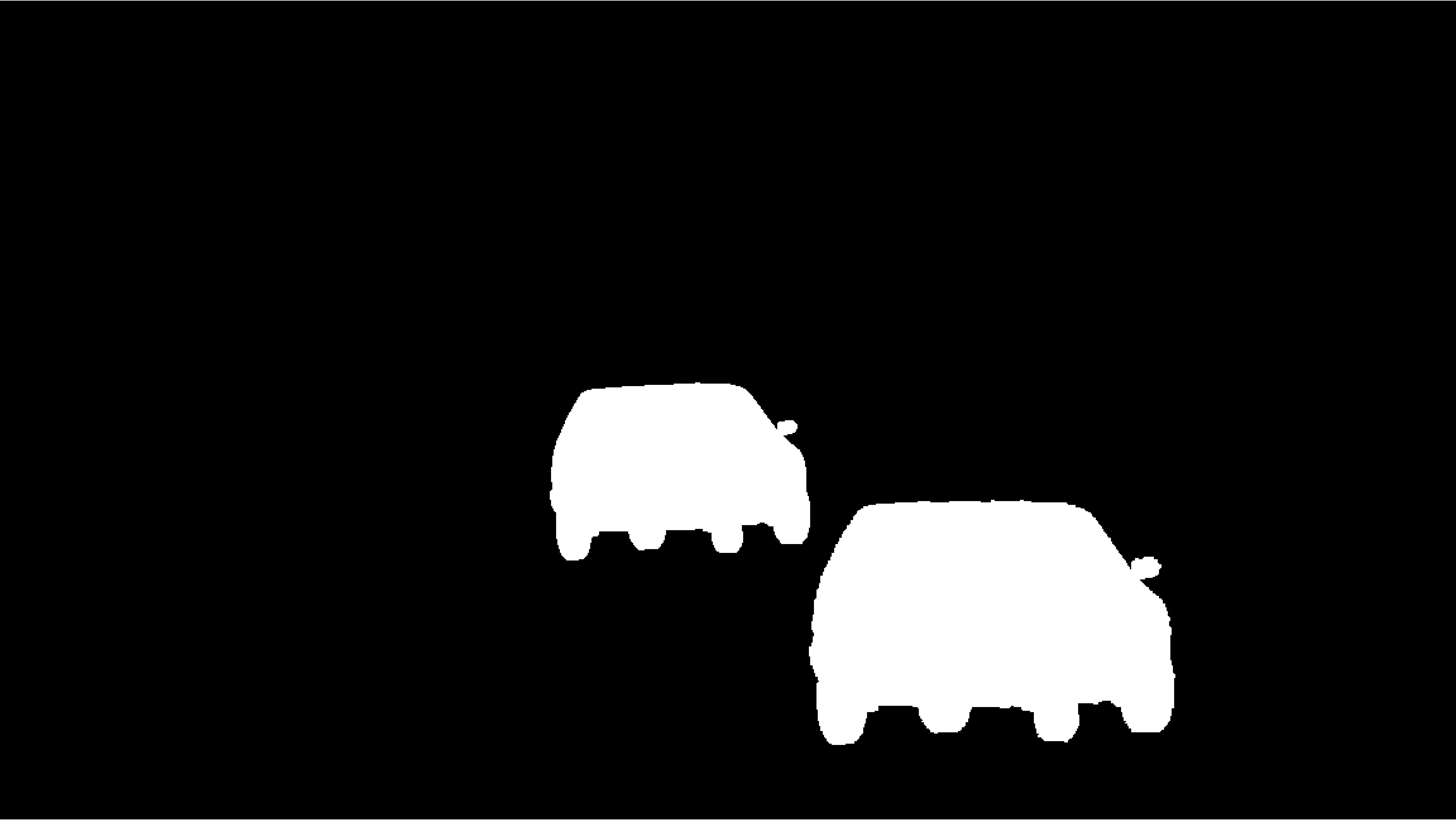} \\
		\includegraphics[width=0.23\linewidth]{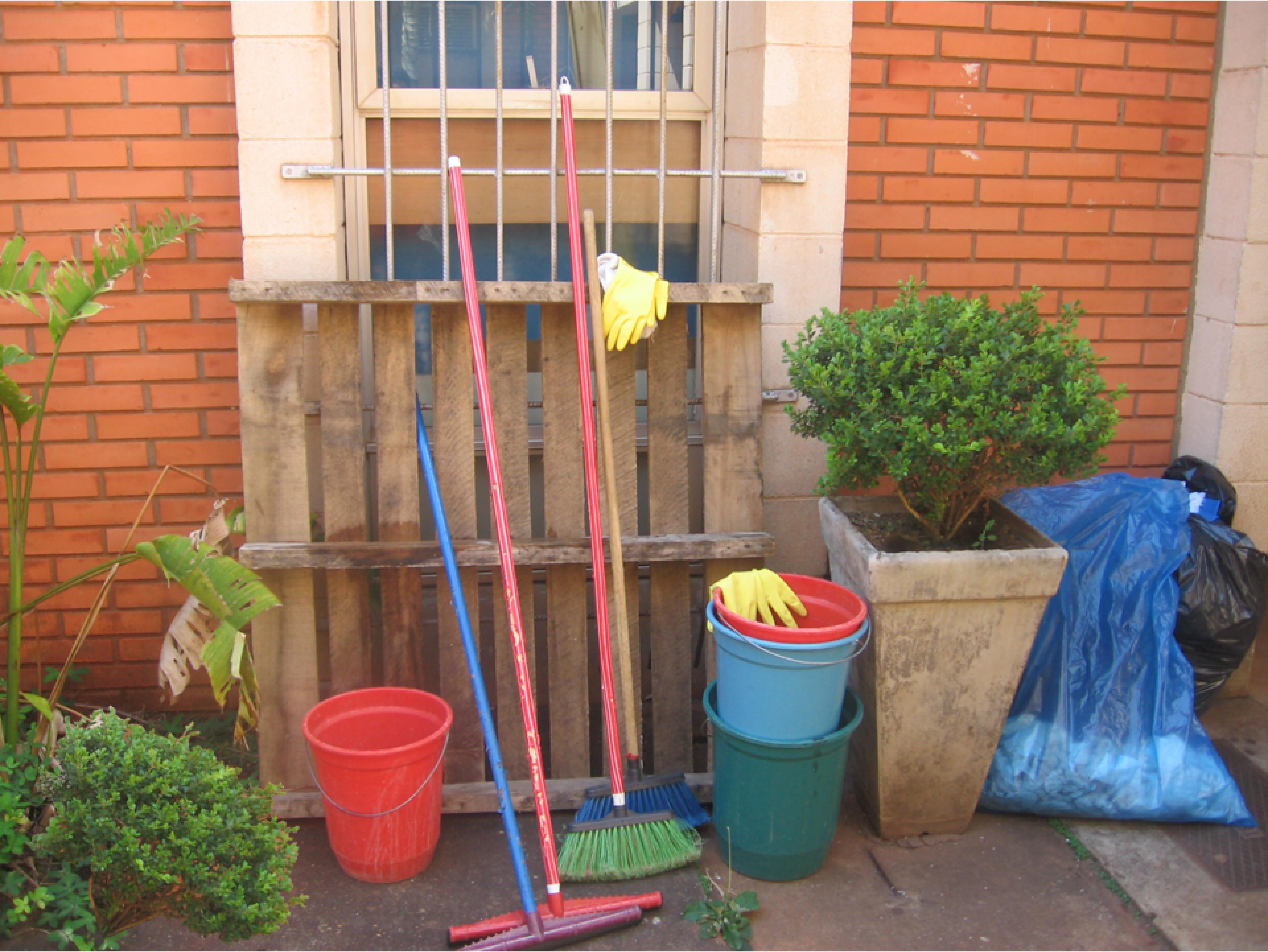} &
		\includegraphics[width=0.23\linewidth]{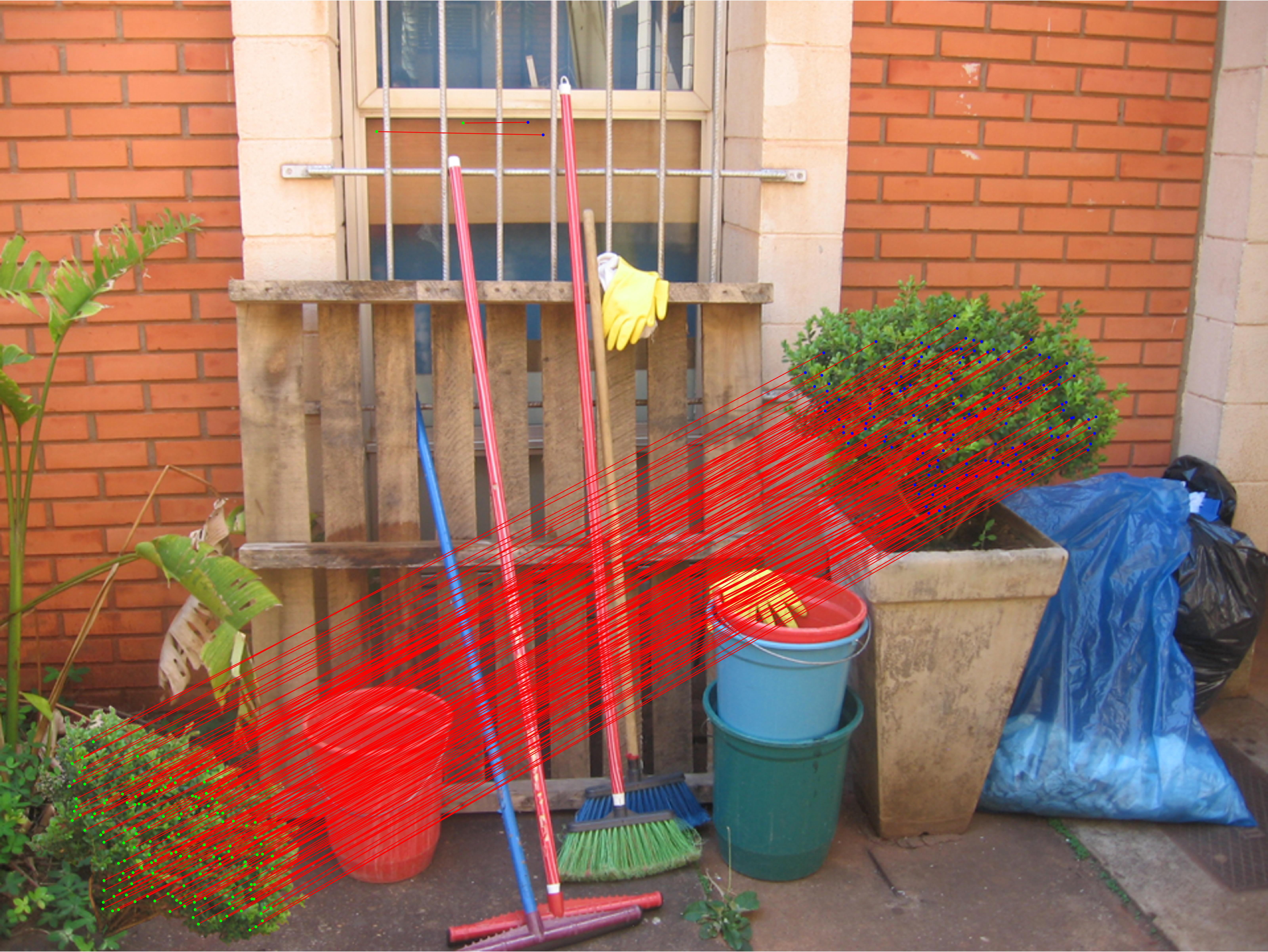} &
		\includegraphics[width=0.23\linewidth]{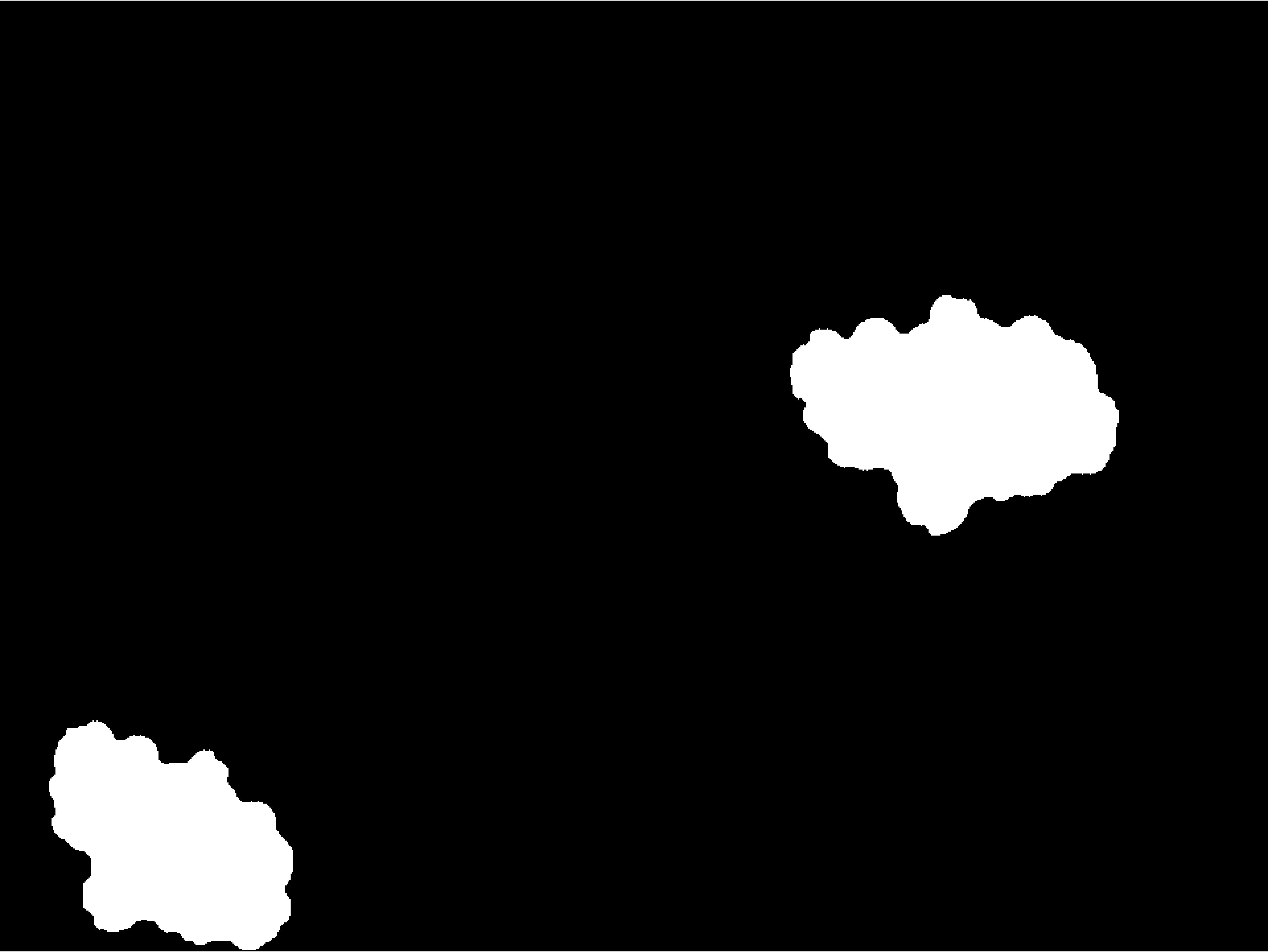} &
		\includegraphics[width=0.23\linewidth]{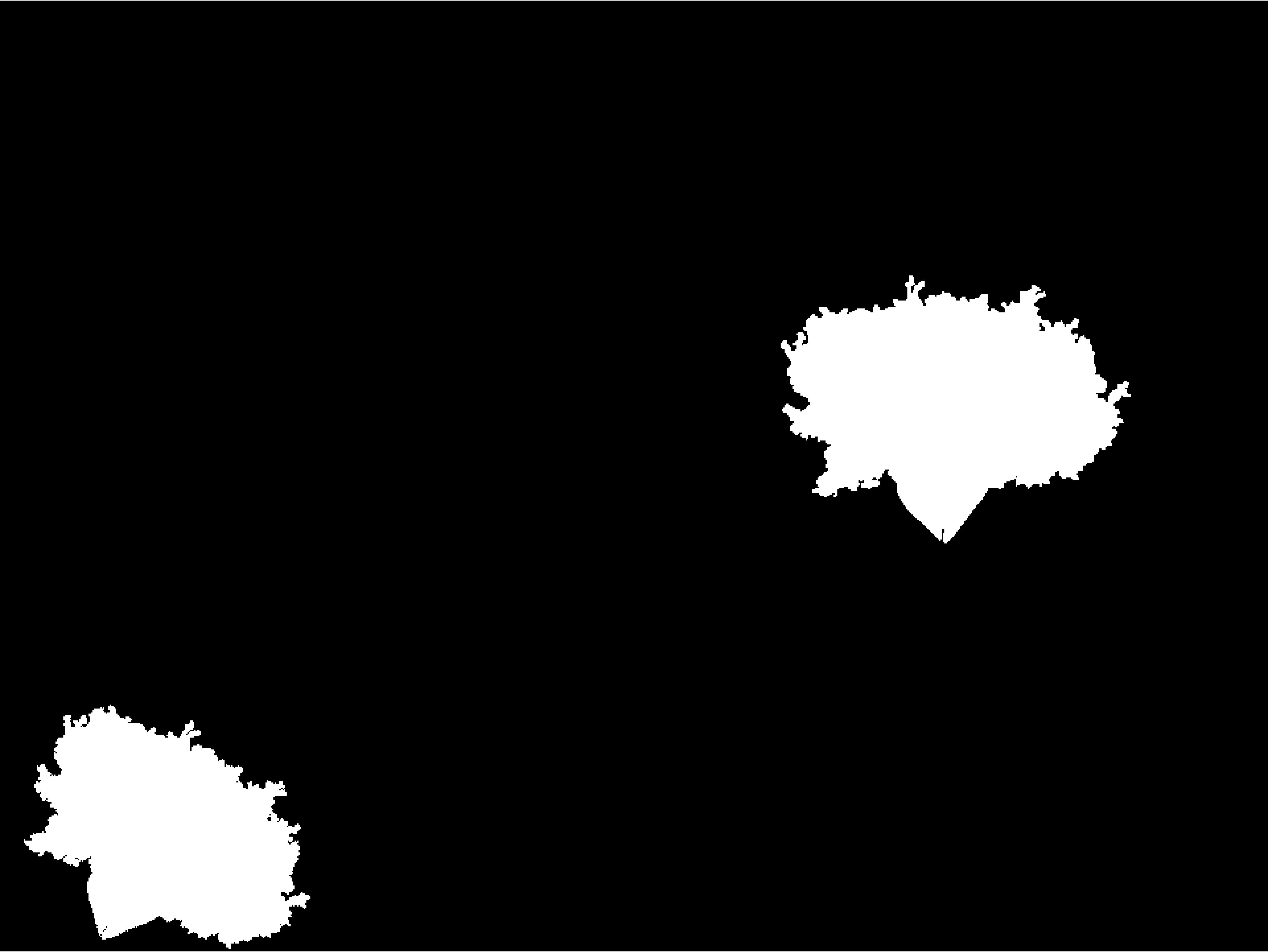} \\
		(a) & (b) & (c) & (d)
	\end{tabular}
	\caption{Some experimental results. (a) forgery images; (b) matching; (c) binary result; (d) ground-true.}
	\label{Fig7}
\end{figure*}

\subsection{Evaluation Metrics}\label{Sec4_2}
Generally, the evaluation of CMFD techniques can be performed at two levels: image level and pixel level. At the image level, the objective is to accurately determine whether an image is tampered with or original. At the pixel level, a more stricter criterion is applied, emphasizing the localization of tampered regions within the image. Usually, datasets commonly employ three evaluation metrics, which are defined as follows:
\begin{equation}
	\label{Eq16}
	\textit{TPR} = \frac{{\textit{TP}}}{{\textit{TP} + \textit{FN}}}
\end{equation}
\begin{equation}
	\label{Eq17}
	\textit{FPR} = \frac{{\textit{FP}}}{{\textit{TN} + \textit{FP}}}
\end{equation}
\begin{equation}
	\label{Eq18}
	\textit{F} = \frac{{2\textit{TP}}}{{2\textit{TP} + \textit{FP} + \textit{FN}}}
\end{equation}

In the equation mentioned above, $ \textit{TP} $ represents the number of correctly detected tampered images or pixels, $ \textit{TN} $ represents the number of correctly detected original images or pixels, $ \textit{FN} $ represents the number of incorrectly detected tampered images or pixels, and $ \textit{FP} $ represents the number of incorrectly detected original images or pixels. To provide more detailed test results, $ \textit{TPR} $ and $ \textit{FPR} $ are used to represent image-level $ \textit{TPR} $ and $ \textit{FPR} $ metrics, while $ \textit{F-i} $ and $ \textit{F-p} $ represent image-level and pixel-level $ \textit{F} $ metrics, respectively.

\subsection{Analysis Of Parameters}\label{Sec4_3}
In this subsection, this paper mainly analyzes the pre-processing resize factor $ s $ and entropy radius $ R_E $, as well as the matching process parameters $ step_3 $ and $ step_4 $.

\begin{figure}[ht]
	\centering
	\begin{tabular}{cc}
		\includegraphics[width=0.48\linewidth]{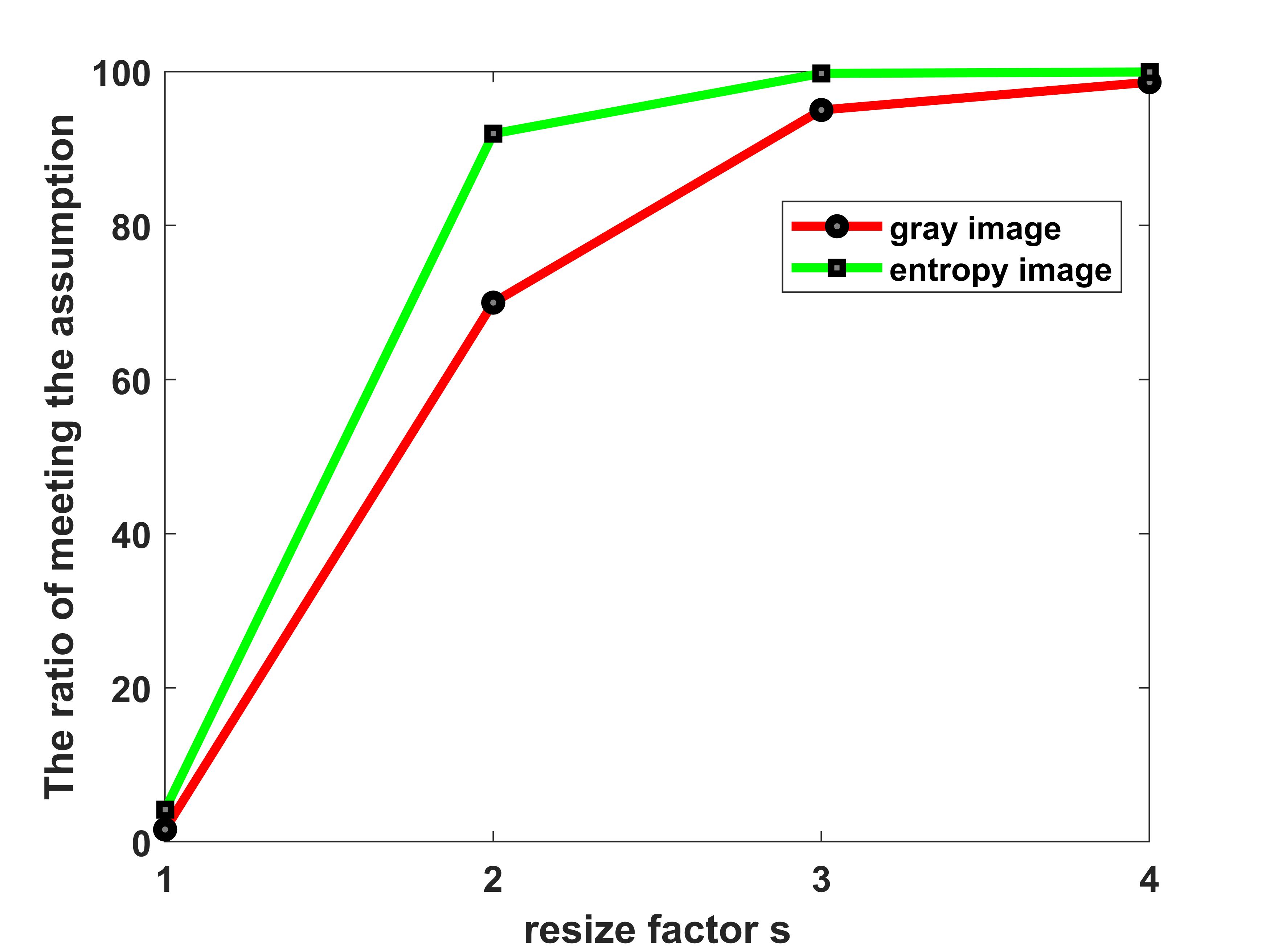} &
		\includegraphics[width=0.48\linewidth]{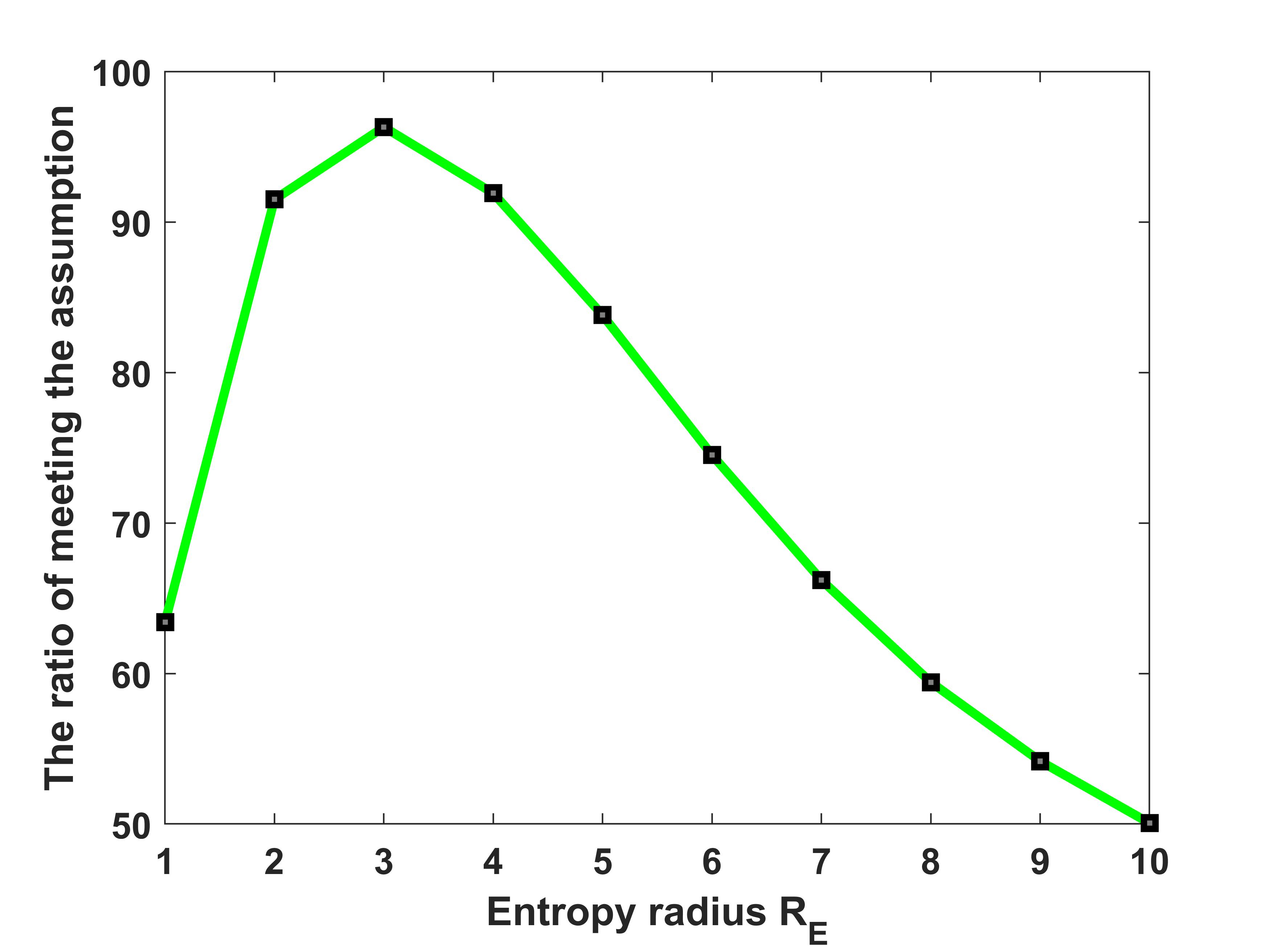} \\
		(a) & (b)
	\end{tabular}
	\caption{The analysis of pre-processing. (a) the ratio of meeting the assumption under different resize factor $ s $; (b) The ratio of meeting the assumption under different entropy radius $ R_E $.}
	\label{Fig5}
\end{figure}

As shown in Fig. \ref{Fig5} (a), this paper calculates the ratio of meeting the assumption (\ref{Sec2_2}) under different $ s $. As resize factor $ s $ increases, more pixel patches have 4 keypoints. Obviously, when the resize factor $ s $ is the equal, there is considerable evidence that the entropy image is more suitable for CMFD. In this paper, we adopt $ s=2 $.

As shown in Fig. \ref{Fig5} (b), this paper calculates the ratio of meeting the assumption under different $ R_E $. Furthermore, the ratio of meeting the assumption in gray images is only 69.99\%. Clearly, entropy images yields better results compared to gray images when $ R_E $ is within the range [2, 6], and the highest performance occurs at $ R_E=3 $. Therefore, this paper adopts $ R_E=3 $.

\begin{figure}[ht]
	\centering
	\begin{tabular}{cc}
		\includegraphics[width=0.48\linewidth]{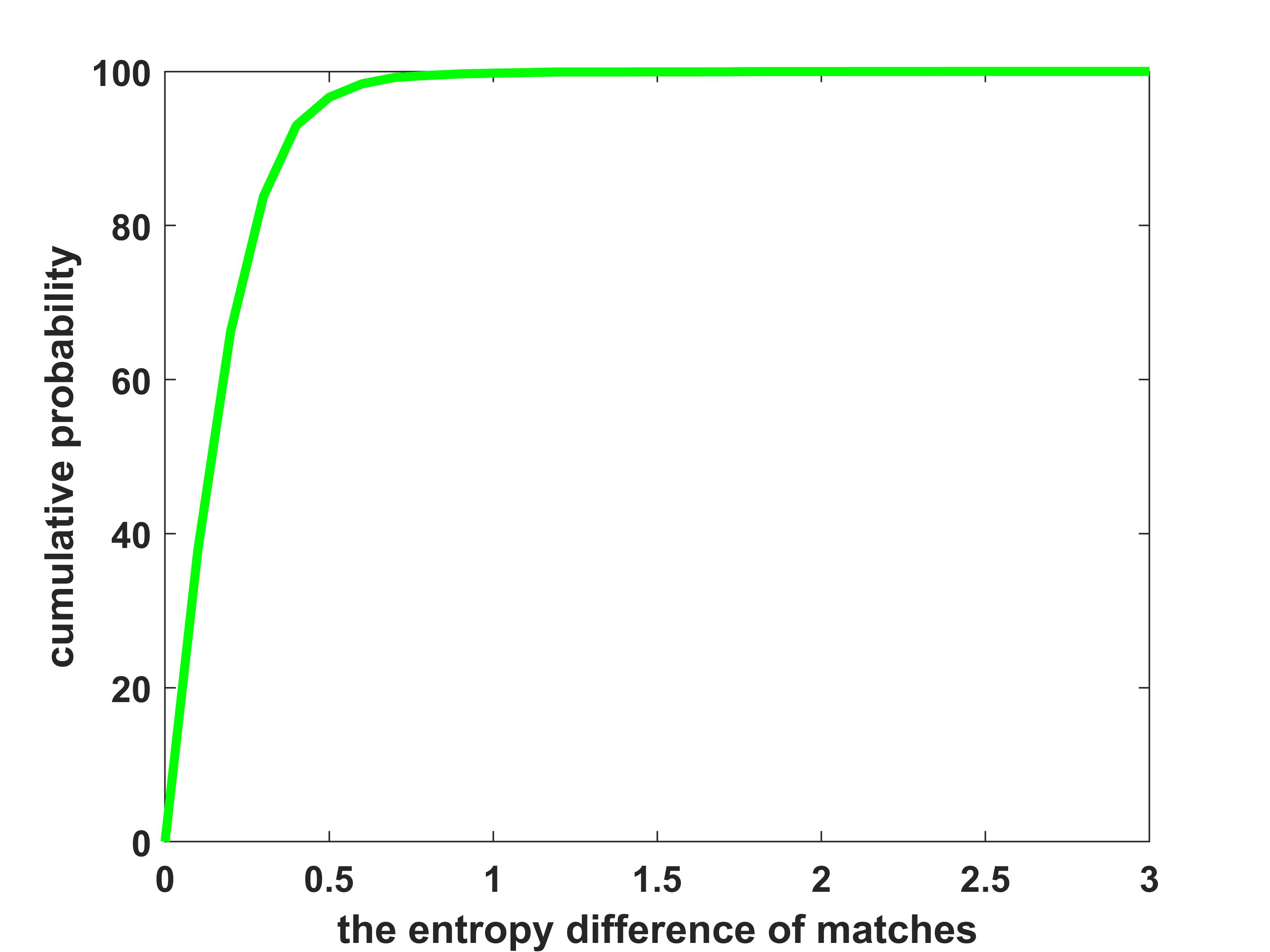} &
		\includegraphics[width=0.48\linewidth]{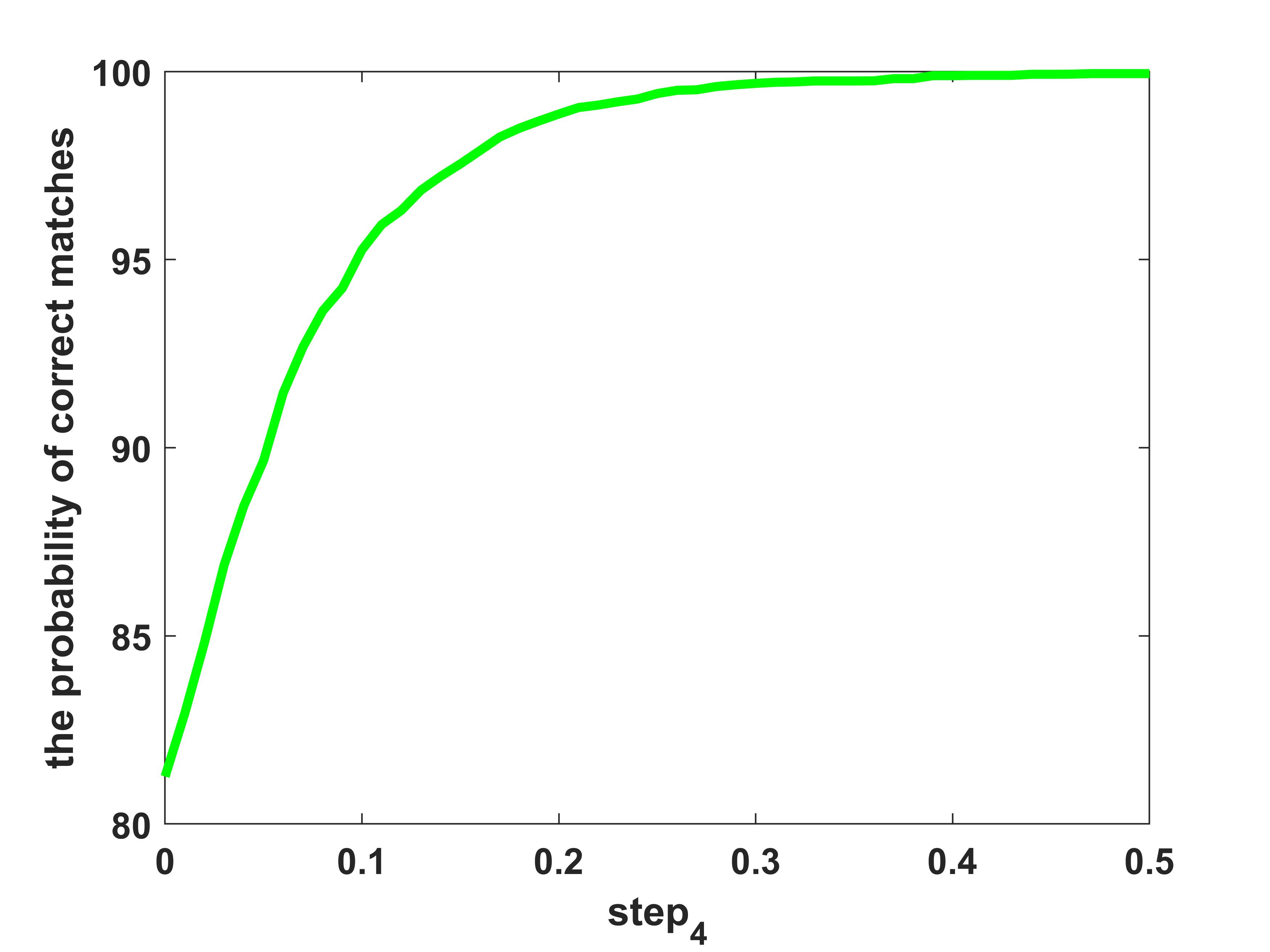} \\
		(a) & (b)
	\end{tabular}
	\caption{The analysis of overlapped entropy level clustering. (a) the cumulative probability of the entropy difference of matches; (b) the number of correct matches relative to brute force matching under different $ step_{4} $.}
	\label{Fig6}
\end{figure}

Considering that the robustness of entropy values may diminish when facing geometric transformation, the fourth group of images from the CMH dataset are used for brute force matching and analyze the distribution of entropy values for correct matches. To quantify $ step_{3} $, this paper calculates the entropy difference between correct matches. Then, the cumulative probability is obtained, as shown in Fig. \ref{Fig6} (a). This paper believes that the entropy difference for correct matches lies in the range of [0, 1] (The cumulative probability reaches 99.77\% at $ step_{3}=1 $). Therefore, this paper chooses $ step_{3}=1 $.

As shown in Fig. \ref{Fig6} (b), statistical analysis is conducted with a step size of 0.01 over the range of [0, 0.5]. Based on observation, the correct matching reached 81.26\% when $ step_{4}=0 $. As $ step_4 $ increases, the rate of correct matches relative to brute force matching continues to rise in the overlapped entropy clustering. Considering that a larger overlapped interval leads to higher matching complexity, this paper adopts $ step_4=0 $ (If readers want to accelerate using ANN, this paper recommends $ step_4=0.2 $, which achieves an accuracy rate of 99.04\%).

\subsection{Performance Comparison On Different Datasets}\label{Sec4_4}
In this work, a combined dataset named CMH+GRIPori is adopted,
consisting of all the 108 forgeries from CMH, and the 80 original images from GRIP dataset. Some results of the testing are shown in Fig. \ref{Fig7}.

Table \ref{Tab2} lists the performance comparison of different algorithms on CMH+GRIPori dataset, including keypoint-based \cite{1,2,3,14}, block-based \cite{6} and the proposed method. Obviously, our proposed method achieves the best results in comparison with the current best classical algorithms.
\begin{table}[htb]
	\centering
	\caption{Performance comparison of different algorithms on the CMH+GRIPori dataset}
	\label{Tab2}
	\begin{tabular}{cccccc} \hline 
		methods & $ \textit{TPR} $ & $ \textit{FPR} $ & $ \textit{F-i} $ & $ \textit{F-p} $  & time  \\ \hline
		\cite{1} & 96.3	&	0	&	98.11	&	90.61	&	$ \textbf{11.2s} $ \\
		\cite{2} & 	95.37	&	38.75	&	85.12	&	64.09	&	18.1s \\
		\cite{3} &	$ \textbf{99.07} $	&	$ \textbf{0} $	&	$ \textbf{99.53} $	&	91.72	&	43.3s \\
		\cite{14}&	$ \textbf{99.07} $	&	11.25	&	95.54	&	91.09	&	22.2s\\
		\cite{6} &	92.59	&	8.75	&	93.02	&	88.10	&	13.1s \\
		proposed & $ \textbf{99.07} $	&	$ \textbf{0} $	&	$ \textbf{99.53} $	&	$ \textbf{92.47} $	&	19.9s	\\ \hline
	\end{tabular}
\end{table}

Table \ref{Tab3} lists the performance comparison of different algorithms on GRIP. It can be observed that our proposed method achieves great results on this dataset as well. However, our proposed method has shortcomings in terms of time complexity, but this is not due to the design of the matching stage. In our statistics, our proposed method obtains approximately 1.5 times the number of keypoints compared to the generated gray images. Typically, the time complexity is directly proportional to the square of the number of keypoints. Based on this estimation, our matching efficiency is approximately 29\% faster than the state-of-the-art method mentioned in reference \cite{3} when tested on the GRIP dataset.

\begin{table}[htb]
	\centering
	\caption{Performance comparison of different algorithms on the GRIP dataset}
	\label{Tab3}
	\begin{tabular}{cccccc} \hline 
		methods & $ \textit{TPR} $ & $ \textit{FPR} $ & $ \textit{F-i} $ & $ \textit{F-p} $  & time  \\ \hline
		\cite{1} & $ \textbf{100} $	&	$ \textbf{0} $	&	$ \textbf{100} $	&	94.66	&	13.9s \\
		\cite{2} &	$ \textbf{100} $	&	38.75	&	83.77	&	66.62	&	$ \textbf{11.7s} $ \\
		\cite{3} &	$ \textbf{100} $	&	$ \textbf{0} $	&	$ \textbf{100} $	&	$ \textbf{98.57} $	&	12.9s \\
		\cite{14}&	$ \textbf{100} $	&	11.25	&	94.67	&	85.71	&	21.5s\\
		\cite{6} & 98.75 &	8.75 & 95.18	& 92.99	 & 14.8s \\
		proposed & $ \textbf{100} $	&	$ \textbf{0} $	&	$ \textbf{100} $	&	95.47	&	20.6s	\\ \hline
	\end{tabular}
\end{table}

\section{Conclusion}\label{Sec5}
In this paper, a novel framework is proposed using entropy information, which can be used for CMFD. Firstly, entropy images are introduced to determine the coordinates and scales space of keypoints. Considering SIFT features represent the gradient information of the grayscale, the keypoints redefine orientation and extraction feature in gray image, which makes our matching process more accuracy. Then, an entropy level clustering is developed to avoid increased matching complexity caused by non-ideal distribution of grayscale in keypoints. Experimental results demonstrate that our algorithm achieves a good balance between performance and time efficiency.

\bibliographystyle{IEEEbib}
\bibliography{References}
\end{document}